\documentclass{article} 
\usepackage{iclr2027_conference,times}

\usepackage{amsmath,amsfonts,bm}

\def\eqref#1{equation~\ref{#1}}

\def\1{\bm{1}}

\DeclareMathAlphabet{\mathsfit}{\encodingdefault}{\sfdefault}{m}{sl}
\SetMathAlphabet{\mathsfit}{bold}{\encodingdefault}{\sfdefault}{bx}{n}

\usepackage{hyperref}
\usepackage{url}
\usepackage{booktabs}
\usepackage[table]{xcolor}
\usepackage{multirow}
\usepackage{array}
\usepackage{amssymb}
\usepackage{tabularx}
\usepackage{graphicx}
\usepackage{caption}
\usepackage{enumitem}
\usepackage{makecell}
\usepackage[most]{tcolorbox}
\usepackage{capt-of} 

\setcitestyle{comma}

\newtcolorbox{promptbox}[1]{
    enhanced,
    breakable,
    colback=gray!3,
    colframe=gray!40,
    colbacktitle=gray!25,
    coltitle=black,
    title=#1,
    fonttitle=\bfseries\normalsize,
    boxrule=0.6pt,
    arc=2mm,
    outer arc=2mm,
    left=8pt,
    right=8pt,
    top=7pt,
    bottom=7pt,
    toptitle=2.5pt,
    bottomtitle=2.5pt
}

\title{PLAN-Bench: Can Existing VLMs Go Well Beyond Perception to Reason and Plan in Embodied Worlds?}

\author{%
\begin{minipage}[t]{\dimexpr\textwidth-2\tabcolsep\relax}
\centering
\normalfont\normalsize
\textbf{Juntong Feng}$^{1*}$,
\textbf{Qingan Liu}$^{1*}$,
\textbf{Jingru Dong}$^{1}$,
\textbf{Yue Zhu}$^{1}$,
\textbf{Weiming Zheng}$^{1}$,
\textbf{Yujie Cheng}$^{1}$\\
\textbf{Xinzhe Han}$^{1}$,
\textbf{Shuai Hao}$^{1}$,
\textbf{Yunzhi Zhuge}$^{1}$,
\textbf{Haiwen Diao}$^{2\dagger}$,
\textbf{Huchuan Lu}$^{1}$\\[0.3em]
$^{1}$Dalian University of Technology,
$^{2}$Nanyang Technological University
\endgraf
\end{minipage}%
}

\iclrfinalcopy

\begin{document}

\maketitle

\pagestyle{fancy}
\fancyhf{}
\fancyfoot[C]{\thepage}
\renewcommand{\headrulewidth}{0pt}
\renewcommand{\footrulewidth}{0pt}
\fancypagestyle{preprintfirstpage}{%
    \fancyhf{}
    \fancyfoot[L]{\footnotesize Preprint.}
    \renewcommand{\headrulewidth}{0pt}
    \renewcommand{\footrulewidth}{0pt}
}
\thispagestyle{preprintfirstpage}

\ificlrfinal
\begingroup
\renewcommand{\thefootnote}{}
\footnotetext[0]{%
$^{*}$Equal contribution.
$^{\dagger}$Correspondence to
\texttt{haiwen.diao@ntu.edu.sg}%
}
\endgroup
\fi

\begin{abstract}

Vision-Language Models (VLMs) have achieved strong performance in visual perception and spatial reasoning, leading to growing interest in using them as high-level planners for embodied AI.
However, physical tasks require models to integrate spatial information, reason over temporal states, and generate multi-step plans under various constraints, raising the question of whether current VLMs can reliably support such reasoning and planning independently of execution.
We introduce \textbf{Plan-Bench}, a benchmark for evaluating the high-level reasoning and multi-step planning capabilities of VLMs relevant to embodied worlds, while isolating the influence of action execution.
It covers three complementary dimensions, spanning 16 subclasses and 699 curated samples.
Extensive evaluation of 28 VLMs reveals room for improvement, with even the strongest model achieving only 75.0\% accuracy.
We further find that model scaling generally improves these capabilities.
Ablation studies also indicate larger models show increasing potential as high-level planners.
Overall, Plan-Bench provides a lightweight probe for evaluating VLMs' reasoning and planning capabilities relevant to embodied environments and assessing their potential as decision makers for physical tasks.
Our benchmark and code are available at
\url{https://github.com/MinneFeng/Plan-Bench}.

\end{abstract}

\section{Introduction}

Recent advances in vision-language models (VLMs)~\citep{openai2026GPT5.6, Qwen2026qwen3.8-max, diao2026sensenova, diao2026sensenova1.5} have substantially strengthened their capabilities in visual perception and spatial reasoning, leading to growing interest in using them as high-level planners for embodied AI~\citep{zhang2026harness, liu2026goal2skill, yang2026hivla}, particularly in complex physical tasks.
In such settings, models often need to integrate spatial information, reason over evolving temporal states, and generate multi-step plans under various constraints.
These demands expose an important and underexplored challenge: can existing VLMs go well beyond perception to reason and plan in embodied worlds?

\begin{figure}[t]
    \centering
    \includegraphics[width=\linewidth]{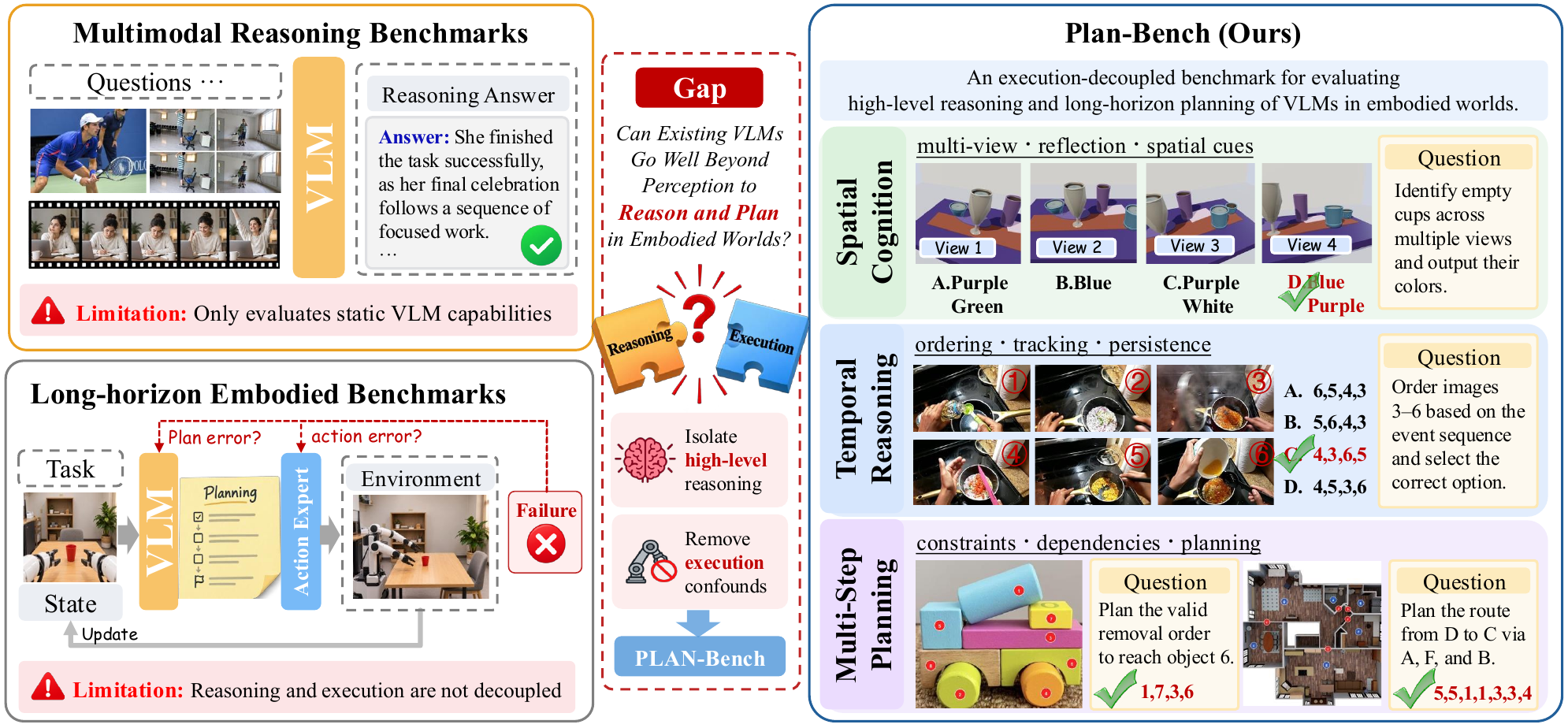}
    \caption{Overview of Plan-Bench. The left illustrates the limitations of existing benchmarks, which either evaluate static VLM capabilities or couple high-level reasoning with low-level execution. The right presents the composition of Plan-Bench, which evaluates execution-decoupled high-level reasoning and planning across Spatial Cognition, Temporal Reasoning, and Multi-Step Planning.}
    \label{fig:motivation}
\end{figure}

Notably, existing benchmarks have not evaluated these capabilities of VLMs for embodied tasks.
As shown in Figure \ref{fig:motivation}, multimodal reasoning benchmarks~\citep{fu2026mme, LENS} primarily focus on visual information.
Embodied benchmarks~\citep{fan2026robospa, han2026robocerebra} mainly evaluate embodied tasks, mixing errors from decisions and execution.
LongAct~\citep{zhu2026robots} attempts to reduce the influence of control and better assess decision making, but evaluates agents rather than VLMs.
It therefore remains unclear whether VLMs can perform reliable reasoning and planning in embodied scenarios when high-level decisions are separated from low-level execution.

To bridge this gap, we introduce Plan-Bench, a benchmark designed to evaluate VLMs' high-level reasoning and multi-step planning relevant to embodied tasks.
As illustrated on the right side of Figure~\ref{fig:motivation}, Plan-Bench systematically covers three dimensions: Spatial Cognition, Temporal Reasoning, and Multi-Step Planning, spanning 16 subclasses and 699 samples.
It is constructed through a three-stage pipeline consisting of task definition, dual-source visual data acquisition, and manual QA construction with quality assurance, resulting in a diverse and carefully curated benchmark dataset.
Plan-Bench includes both single-choice and open-ended questions, and provides answer hints for samples with multiple valid answers to guide more reliable LLM-as-judge evaluation.

We evaluate Plan-Bench on 28 VLMs.
The results show that current VLMs remain limited in reasoning and planning for embodied scenarios.
The best-performing model, Qwen3.8-Max~\citep{Qwen2026qwen3.8-max}, achieves only 75.0\%.
We further observe that performance generally improves with model scaling, with some variation across model families.
Our ablation studies show that larger models exhibit greater potential as high-level planners.
Overall, Plan-Bench provides a principled testbed for evaluating reasoning and planning in VLMs and a lightweight framework for assessing their potential as high-level embodied planners.
Our main contributions are summarized as follows:
\begin{itemize}[itemsep=4pt, topsep=3pt, parsep=0pt, partopsep=0pt]
    \item \textbf{Designing a reasoning and planning benchmark for VLMs in embodied worlds.} Plan-Bench is an easy-to-deploy benchmark spanning 16 subclasses and 699 samples, enabling direct evaluation of VLMs' high-level capabilities independently of low-level execution.
    \item \textbf{Providing a comprehensive evaluation of existing VLMs.} We benchmark 28 representative VLMs and reveal their capabilities in high-level reasoning and multi-step planning.
    \item \textbf{Revealing current limitations and potential.} Evaluation across proprietary and open-source models shows that reasoning and planning in embodied scenarios remain challenging, while larger models exhibit greater potential as high-level planners for embodied tasks.
\end{itemize}

\section{Related Work}

\subsection{Multimodal Reasoning Benchmarks}

Existing benchmarks have gradually evolved from perception toward complex reasoning.
MME~\citep{fu2026mme} evaluates both perceptual and cognitive abilities, while LENS~\citep{LENS} organizes multimodal evaluation from  understanding to reasoning.
Recent benchmarks such as VisuLogic~\citep{xu2026visulogic}, MIRA~\citep{MIRA}, and PuzzleWorld~\citep{PuzzleWorld} further explore chain-of-thought and open-ended multi-step reasoning.
Beyond single-image reasoning, MuirBench~\citep{wang2025muirbench}, MMIU~\citep{meng2025mmiu}, EgoPlan-Bench2~\citep{qiu2026egoplan}, among others~\citep{li2026mmr, meng2025mmiu} extend evaluation to multiple images or frames, covering temporal changes and long-range visual evidence.
These benchmarks broaden the scope and complexity of multimodal reasoning.
However, existing benchmarks rarely target the reasoning and planning demands of embodied worlds.
Plan-Bench addresses this gap by evaluating these capabilities across Spatial Cognition, Temporal Reasoning, and Multi-step Planning.

\subsection{Multi-Step Tasks in Embodied Environments}

Recent embodied systems increasingly employ VLMs~\citep{diao2024unveiling, diao2502evev2} as decision makers.
Harness VLA~\citep{zhang2026harness}, Goal2Skill~\citep{liu2026goal2skill}, and HiVLA~\citep{yang2026hivla}, among others~\citep{shi2025hi,li2025hamster,yang2025guiding}, adopt hierarchical frameworks that separate high-level reasoning and planning from low-level execution.
Meanwhile, several benchmarks have begun to evaluate these capabilities in embodied settings.
EmbodiedBench~\citep{yang2025embodiedbench} and RoboCerebra~\citep{han2026robocerebra} evaluate embodied agents through interactive tasks or hierarchical VLM-VLA systems, where performance jointly reflects reasoning, grounding, and action execution.
RoboSPA~\citep{fan2026robospa} further evaluates VLA models on reasoning and procedural planning, but still couples reasoning with low-level execution.
LongAct~\citep{zhu2026robots} attempts to reduce the influence of control to better assess planning, yet evaluates complete embodied agents rather than VLMs alone.
Plan-Bench instead decouples embodied action execution and directly evaluates VLMs' reasoning and planning by constructing a benchmark of 699 questions, providing a lightweight interface for isolating the capabilities required by embodied tasks.

\section{Plan-Bench}
\label{headings}

We introduce Plan-Bench, a benchmark designed to systematically evaluate the reasoning and planning capabilities of VLMs in embodied tasks.
In this section, we first describe the construction pipeline, then present the details of data distribution, and finally introduce the evaluation protocol.

\begin{figure}[t]
    \centering
    \includegraphics[width=\linewidth]{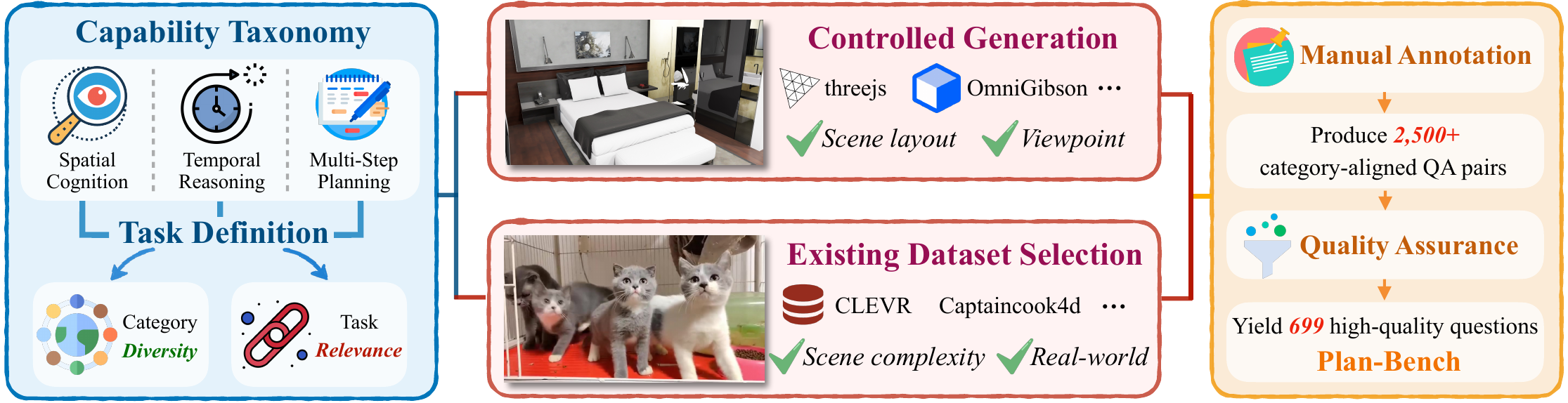}
    \caption{Overview of the Plan-Bench construction pipeline, including capability-driven task definition, dual-source data acquisition, manual annotation, and quality assurance.}
    \label{fig:construction_pipeline}
\end{figure}

\subsection{Construction Pipeline}

To ensure high quality and validity, we adopt a task-driven, multi-stage construction pipeline, as illustrated in Figure \ref{fig:construction_pipeline}. The overall process consists of three stages: 

\textbf{Capability Taxonomy and Task Definition.}
We first organize the capabilities required for reasoning and planning in embodied environments into a taxonomy.
The taxonomy comprises three core categories corresponding to Spatial Cognition, Temporal Reasoning, and Multi-Step Planning.
Based on this taxonomy, we further define a set of subtasks, balancing category diversity with task relevance to the targeted capabilities.
This step provides explicit guidance for data acquisition, determining what types of visual evidence and task conditions should be collected or constructed.

\textbf{Dual-Source Data Acquisition.}
We collect visual data from two sources.
For subclasses requiring precise control over factors such as scene layout and viewpoint, we employ controlled-generation platforms such as Three.js~\citep{danchilla2012three} and OmniGibson~\citep{li2023behavior} to configure task-relevant variables for consistent construction.
For those that benefit from more natural visual distributions, we select suitable images from existing datasets such as CLEVR~\citep{johnson2017clevr} and Captaincook4d~\citep{peddi2024captaincook4d}, among others~\citep{yang2019my,dai2026robomme,ding2023mevis,lin2020progressive}, to increase visual variety and capture greater real-world scene complexity.

\textbf{QA Annotation and Quality Assurance.}
Based on the acquired images, we manually construct 2,500+ questions and answers according to the task specification of each subclass.
All samples are then independently reviewed by three annotators who are not involved in the initial annotation process.
The review considers answer correctness, image-QA consistency, and relevance to the corresponding task definition.
A sample is accepted only when three reviewers agree that it satisfies all criteria.
Finally, this process yields 699 high-quality questions to construct Plan-Bench.

\begin{figure}[!t]
    \centering
    \includegraphics[width=\linewidth]{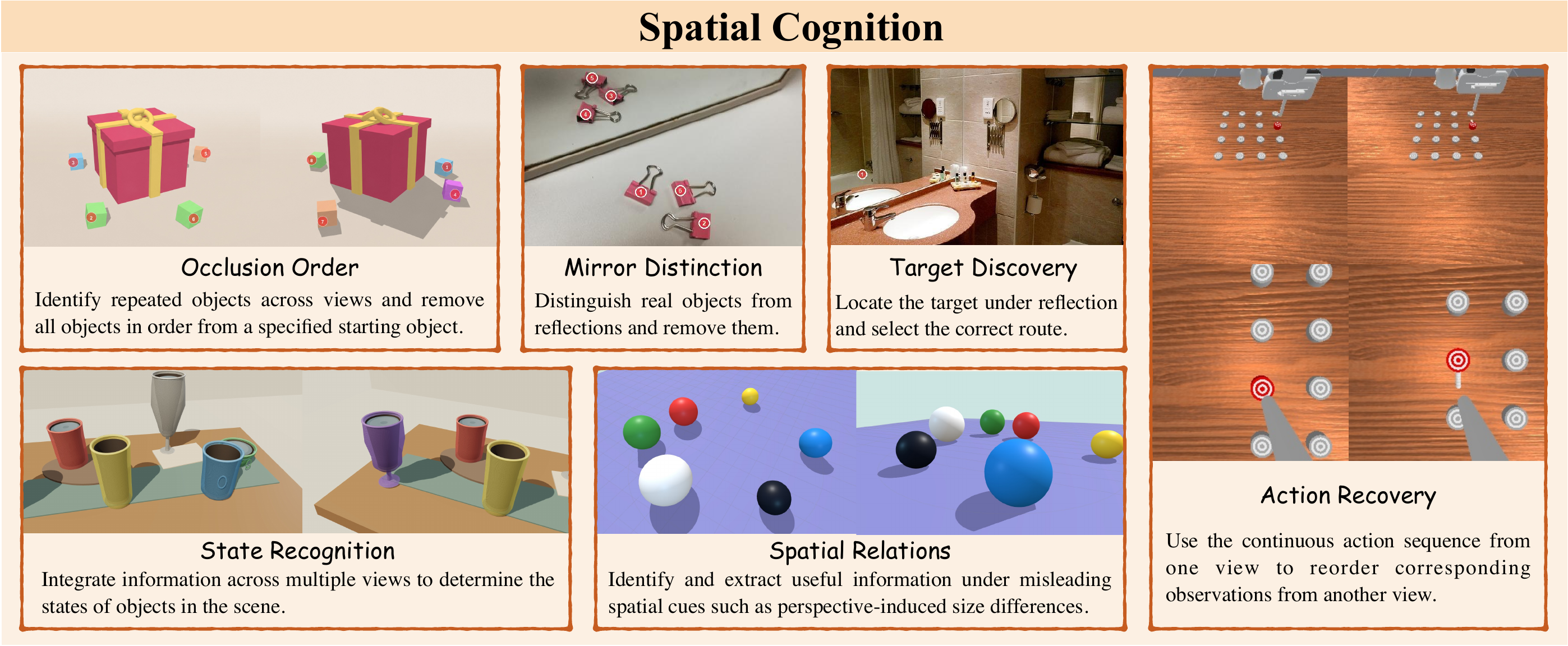}
    \vspace{-3.4mm}

    \includegraphics[width=\linewidth]{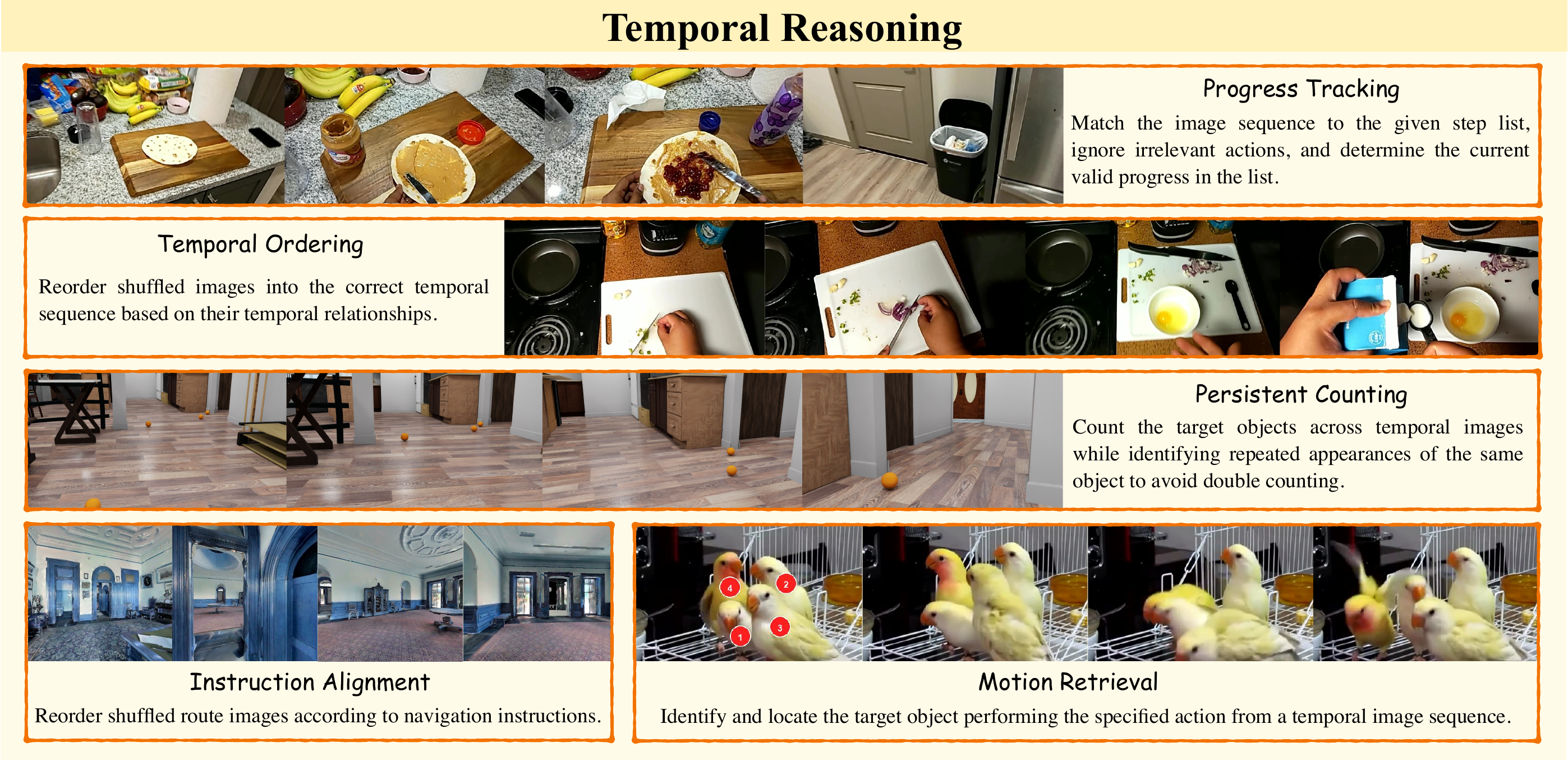}
    \vspace{-3.4mm}

    \includegraphics[width=\linewidth]{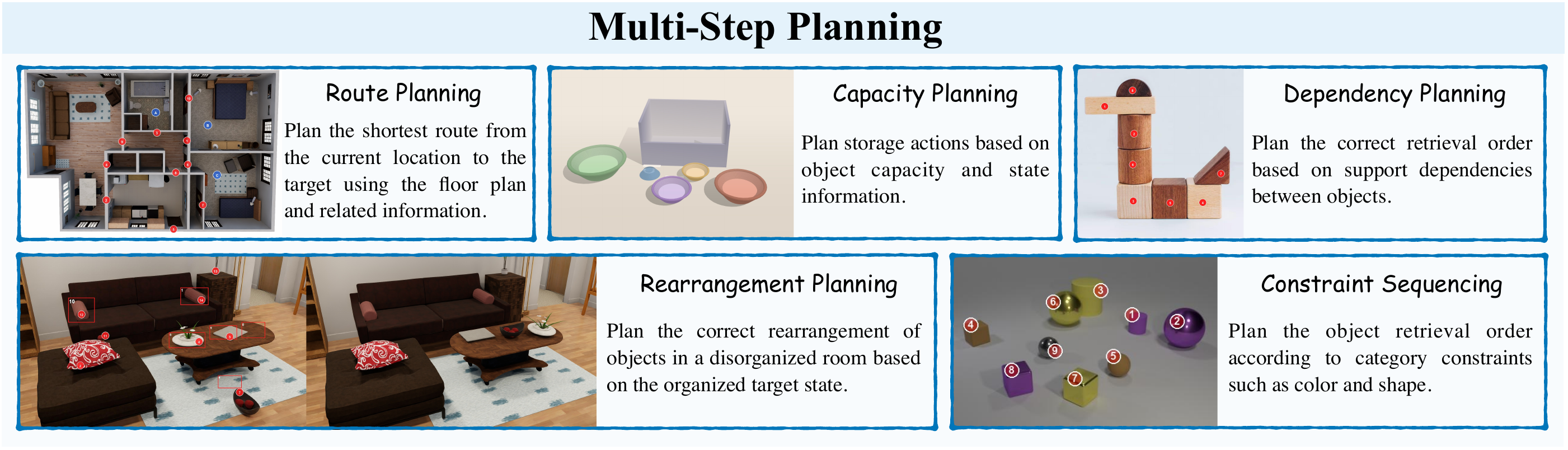}
    \caption{Illustration of representative samples from the three dimensions of Plan-Bench, including Spatial Cognition, Temporal Reasoning, and Multi-Step Planning.}
    \label{fig:category_overview}
\end{figure}

\subsection{Data Distribution}
Plan-Bench comprises 699 instances, including 183 single-choice questions and 516 open-ended questions. 
Our benchmark covers 16 subtasks across three primary categories, providing a comprehensive evaluation of VLMs' reasoning and planning capabilities in diverse embodied task settings.
Figure~\ref{fig:category_overview} presents representative examples from each category in Plan-Bench.
The design motivations and task compositions of these three categories are detailed below:

\textbf{Spatial Cognition.}
This category contains 259 instances across six subtasks: \textit{Occlusion Order}, \textit{Mirror Distinction}, \textit{Action Recovery}, \textit{Target Discovery}, \textit{State Recognition}, and \textit{Spatial Relations}.
It is designed to reflect embodied settings where VLMs may observe the same scene through multiple cameras or viewpoints and must integrate these observations to form a coherent understanding of the environment.
It also considers challenging spatial cues such as mirrors and reflections, requiring VLMs to distinguish reflected content from physical entities and reason correctly about the scene.

\textbf{Temporal Reasoning.}
This category contains 210 instances across five subtasks: \textit{Progress Tracking}, \textit{Temporal Ordering}, \textit{Persistent Counting}, \textit{Instruction Alignment}, and \textit{Motion Retrieval}.
This category includes many egocentric multi-image sequences, designed to reflect embodied tasks such as cooking and navigation, where VLMs must reason over temporally evolving observations from a first-person perspective.
It evaluates whether VLMs can understand temporal dependencies across sequential observations, including ordering events, tracking target actions or states, and identifying recurring objects across frames.
These tasks further assess whether models can preserve relevant historical context and maintain consistent reasoning throughout extended embodied interactions.

\textbf{Multi-Step Planning.}
This category contains 230 instances across five subtasks: \textit{Constraint Sequencing}, \textit{Capacity Planning}, \textit{Rearrangement Planning}, \textit{Dependency Planning}, and \textit{Route Planning}. 
It reflects embodied scenarios where VLMs must first reason about task goals, scene constraints, and action dependencies to produce plans before execution.
It evaluates whether VLMs can translate such information into action sequences toward a target.
Successful planning may indicate the potential of VLMs to serve as more stable high-level decision makers in physical systems.

\definecolor{myblue}{RGB}{235,245,255}

\newcommand{\rrw}[5]{%
  #1 & #2 & #3 & #4 & #5 \\
}
\newcommand{\grp}[1]{%
  \rowcolor{myblue}
  \multicolumn{5}{l}{\textbf{$\blacktriangledown$ #1}} \\
  \noalign{\vskip 1.0pt}
}

\begin{table*}[!t]

\caption{
Comparison of different VLMs on Plan-Bench.
We report performance across the three task categories: Spatial Cognition, Temporal Reasoning, and Multi-Step Planning, together with the overall accuracy.
The \textbf{best} results in each column are bolded.
The random baseline is computed by assigning 0\% to open-ended questions and 25\% to single-choice questions.
All values are rounded to the nearest tenth of a percentage point and reported as percentages (\%).
}
\label{tab:main_results}

\centering
\begin{small}

\setlength{\tabcolsep}{2.5pt}
\renewcommand{\arraystretch}{1.08}

\begin{tabular}{@{}
>{\raggedright\arraybackslash}p{0.4\textwidth}|
>{\centering\arraybackslash}p{0.138\textwidth}|
>{\centering\arraybackslash}p{0.138\textwidth}|
>{\centering\arraybackslash}p{0.138\textwidth}|
>{\centering\arraybackslash}p{0.10\textwidth}
@{}}
\toprule

\multirow{2}{*}{\raggedright\textbf{Model}}
& \textbf{Spatial}
& \textbf{Temporal}
& \textbf{Multi-Step}
& \textbf{Overall}
\\[-1pt]

& \textbf{Cognition}
& \textbf{Reasoning}
& \textbf{Planning}
& \textbf{(699)}
\\

\midrule

\rrw{Random Baseline}{10.4}{4.2}{4.3}{6.5}

\midrule

\grp{Commercial Models}

\rrw{Qwen3.8-Max~\citep{Qwen2026qwen3.8-max}}
{\textbf{66.4}}{78.1}{\textbf{81.7}}{\textbf{75.0}}

\rrw{Kimi-K3~\citep{team2026kimi}}
{69.5}{\textbf{78.6}}{63.5}{70.2}

\rrw{GPT-5.6-sol~\citep{openai2026GPT5.6}}
{56.4}{69.5}{73.0}{65.8}

\rrw{Claude-Opus-4.8~\citep{anthropic2026claudeopus48}}
{42.5}{49.1}{63.5}{51.4}

\rrw{Gemini-3.1-Pro-Preview~\citep{google2026gemini31pro}}
{39.8}{57.1}{43.9}{46.4}

\midrule

\grp{Large Open-Weight Models}

\rrw{Qwen3.8-27B~\citep{qwen2026qwen38}}
{\textbf{48.7}}{\textbf{65.2}}{\textbf{74.8}}{\textbf{62.2}}

\rrw{Qwen3.5-27B~\citep{qwen2026qwen35}}
{39.0}{65.7}{66.1}{55.9}

\rrw{Qwen3.5-35B-A3B~\citep{qwen2026qwen35}}
{34.8}{62.4}{54.4}{49.5}

\rrw{Gemma-4-31B-it~\citep{team2026gemma}}
{30.5}{51.0}{51.3}{43.5}

\rrw{Qwen3-VL-30B-A3B~\citep{yang2025qwen3}}
{22.4}{21.9}{15.7}{20.0}

\rrw{InternVL3.5-30B-A3B~\citep{wang2025internvl3}}
{14.7}{25.2}{13.9}{17.6}

\midrule

\grp{Medium Open-Weight Models}

\rrw{Qwen3.5-9B~\citep{qwen2026qwen35}}
{\textbf{29.7}}{\textbf{59.5}}{\textbf{61.7}}{\textbf{49.2}}

\rrw{Step3-VL-10B~\citep{huang2026step3}}
{25.1}{43.8}{40.9}{35.9}

\rrw{Gemma-4-12B-it~\citep{team2026gemma}}
{24.7}{31.9}{29.6}{28.5}

\rrw{GLM-4.6V-Flash~\citep{hong2025glm}}
{23.9}{35.2}{23.0}{27.0}

\rrw{InternVL3.5-8B~\citep{wang2025internvl3}}
{19.3}{28.1}{12.2}{19.6}

\rrw{Ovis2.5-9B~\citep{lu2025ovis2}}
{20.9}{24.3}{13.9}{19.6}

\rrw{NEO-9B-SFT~\citep{diao2026neo}}
{23.6}{19.1}{12.6}{18.6}

\rrw{MiMo-VL-7B-SFT~\citep{yue2025mimo}}
{19.7}{21.4}{13.9}{18.3}

\rrw{SenseNova-U1~\citep{diao2026sensenova}}
{17.4}{15.2}{20.9}{17.9}

\rrw{Qwen3-VL-8B-Instruct~\citep{yang2025qwen3}}
{17.8}{19.5}{16.5}{17.9}

\rrw{NEO-ov~\citep{diao2026neo-ov}}
{18.5}{17.6}{12.6}{16.3}

\rrw{Ovis2-8B~\citep{lu2025ovis2}}
{15.1}{19.1}{13.0}{15.6}

\midrule

\grp{Small Open-Weight Models}

\rrw{Qwen3.5-4B~\citep{qwen2026qwen35}}
{\textbf{27.0}}{\textbf{52.4}}{\textbf{53.0}}{\textbf{43.2}}

\rrw{Qwen3.5-2B~\citep{qwen2026qwen35}}
{15.1}{24.3}{22.2}{20.2}

\rrw{Qwen3-VL-2B-Instruct~\citep{yang2025qwen3}}
{14.3}{17.1}{8.3}{13.2}

\rrw{InternVL3.5-2B~\citep{wang2025internvl3}}
{13.9}{18.6}{5.7}{12.6}

\rrw{NEO-2B-SFT~\citep{diao2026neo}}
{10.4}{7.1}{7.0}{8.3}

\bottomrule
\end{tabular}

\end{small}
\vskip -0.1in
\end{table*}

\subsection{Evaluation Protocol}

We evaluate model responses using an LLM-as-judge approach. 
Many open-ended questions in Plan-Bench admit multiple valid answers due to equivalent object references, alternative orderings, or other semantically equivalent forms.
To handle such cases, we pair these samples with answer hints that specify additional valid alternatives or equivalence conditions beyond the ground-truth answer, such as valid order swaps, position exchanges, or alternative object names.
DeepSeek-V4-Flash~\citep{xu2026deepseek} serves as the judge, taking the model output, ground-truth answer, and answer hint as input to determine whether the response should be considered correct under these acceptable alternatives.
To validate the reliability of the evaluation process, we randomly sample 200 instances for manual verification, and the judge achieves 100\% agreement with human evaluation on these samples.
The full judge prompt is provided in the Appendix~\ref{app:judge_prompt} for reproducibility.

\section{Experiments}
\label{others}

In this section, we present an evaluation of VLMs on Plan-Bench to examine their reasoning and planning capabilities in embodied worlds.
We first introduce the experimental setup, then report the main results on 28 VLMs, and finally provide ablation studies to further investigate model behaviors.

\subsection{Setup}

\textbf{Models.}
We evaluate Plan-Bench on a broad suite of 28 VLMs, including 23 open-source models and 5 commercial models.
The evaluated models are generally grouped into four categories according to model scale and deployment setting: \textit{(i)} lightweight models with around 2B-4B parameters, \textit{(ii)} middle-scale models with 7B--12B parameters, \textit{(iii)} larger open-source models, comprising 27B--31B dense models and Mixture-of-Experts (MoE) models, and \textit{(iv)} commercial frontier models.
For open-source models, we evaluate a diverse set of model families with different architectures, including Qwen3.8~\citep{qwen2026qwen38}, Qwen3.5~\citep{qwen2026qwen35}, Qwen3-VL~\citep{yang2025qwen3}, Gemma-4~\citep{team2026gemma}, Step3-VL~\citep{huang2026step3}, GLM-4.6V~\citep{hong2025glm}, InternVL3.5~\citep{wang2025internvl3}, Ovis~\citep{lu2025ovis2}, SenseNova-U1~\citep{diao2026sensenova}, NEO~\citep{diao2026neo}, and MiMo-VL~\citep{yue2025mimo}.
For commercial models, we evaluate GPT-5.6-sol~\citep{openai2026GPT5.6}, Kimi-K3~\citep{team2026kimi}, Gemini-3.1-pro-preview~\citep{google2026gemini31pro}, Claude-Opus-4.8~\citep{anthropic2026claudeopus48}, and Qwen3.8-Max~\citep{Qwen2026qwen3.8-max}.

\textbf{Evaluation.}
All open-source models are evaluated using VLMEvalKit ~\citep{duan2024vlmevalkit} with the default decoding settings from their official repositories, running in BF16 precision on NVIDIA A800 GPUs.
Proprietary models are evaluated through their official APIs under the same protocol.
Further details on the evaluation prompts and model configurations are provided in the Appendix.

\begin{figure}[t]
    \centering
    \includegraphics[width=\linewidth]{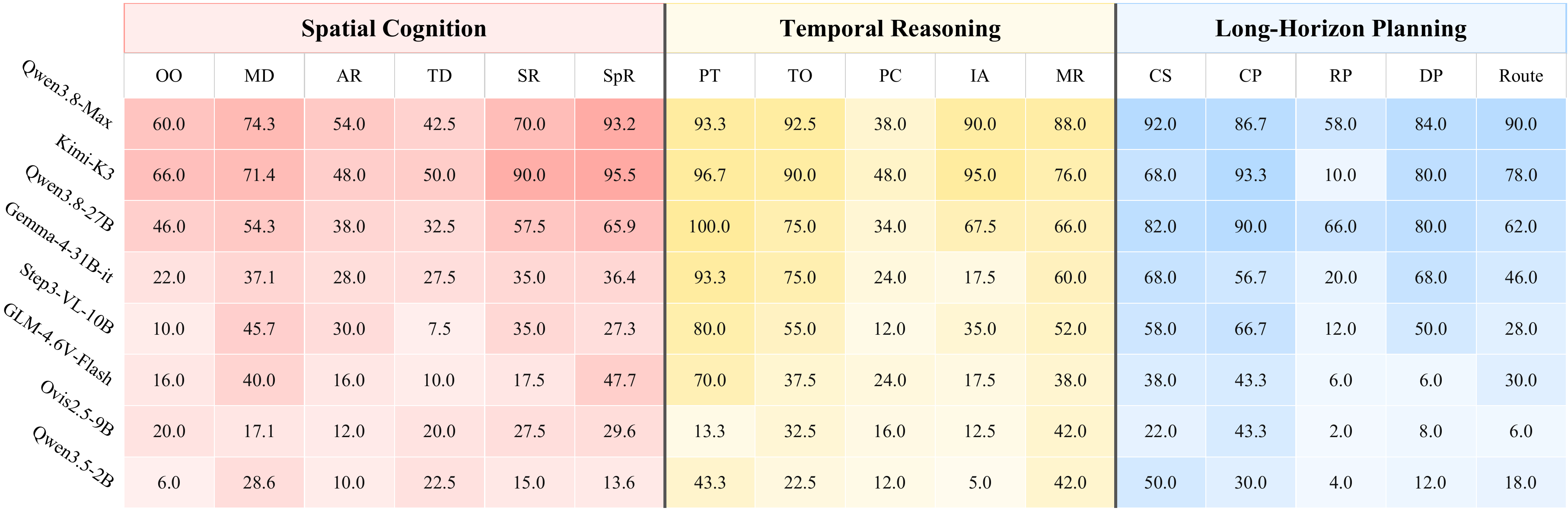}
    \caption{Fine-grained performance of 8 representative VLMs across the 16 subtasks in Plan-Bench. Each cell reports accuracy (\%), with color intensity indicating performance.}
    \label{fig:subtask_headmap}
\end{figure}

\subsection{Main Results}

\textbf{Current VLMs show limited reasoning and planning capabilities toward embodied tasks.}
As shown in Table~\ref{tab:main_results}, current VLMs exhibit room for improvement on Plan-Bench.
Among the 28 evaluated models, Qwen3.8-Max achieves the highest overall accuracy of 75.0\%, followed by Kimi-K3 at 70.2\%. 
All remaining models fall below 70\%, including GPT-5.6-sol at 65.8\%.
Among the MoE, 7--12B, and 2--4B groups, the best-performing models are Qwen3.5-35B-A3B, Qwen3.5-9B, and Qwen3.5-4B, achieving 49.5\%, 49.2\%, and 43.2\% overall accuracy, respectively.
Beyond these leading models, most of the remaining models in these groups cluster around 20\% accuracy, with only a few reaching the high-20\% to mid-30\% range.
These results indicate that reliable reasoning and planning over physical environments remains far from solved for current VLMs.

\textbf{Spatial cognition remains challenging for current VLMs.}
As shown in Table \ref{tab:main_results}, many models stay close to the 10.4\% random baseline.
For example, Qwen3-VL-2B-Instruct and InternVL3.5-2B achieve only 14.3\% and 13.9\%, while Qwen3.5-2B scores 15.1\%.
This weakness is not confined to the lightweight models, as Qwen3-VL-8B-Instruct and NEO-ov reach 17.8\% and 18.5\%.
Across the full set of evaluated models, only nine models exceed 30\% accuracy, while ten remain around or below 20\%.
Even among stronger frontier models, the highest accuracy is only 69.5\%, achieved by Kimi-K3, followed by Qwen3.8-Max at 66.4\%.
These results show that current VLMs still struggle to reliably integrate spatial evidence across multiple viewpoints and complex scene configurations.

\textbf{Model scaling generally benefits embodied reasoning and planning.}
Within the same model family and architecture, performance improves with scale.
Qwen3.5, for example, increases from 20.2\% at 2B to 49.2\% at 9B and further to 55.9\% at 27B.
Similar trends are observed for Gemma-4 and InternVL3.5.
Gemma-4 improves from 28.5\% at 12B to 43.5\% at 31B, while InternVL3.5 rises from 12.6\% at 2B to 19.6\% at 8B.
However, this relationship is not consistent across model families.
Qwen3.8-27B reaches 62.2\%, higher than Gemini-3.1-Pro-Preview at 46.4\%.
Similarly, Qwen3.5-9B achieves 49.2\%, outperforming the much larger Gemma-4-31B at 43.5\%.
Overall, larger models tend to perform better, while architecture and training strategy also remain important factors.

\subsection{Ablation Study}

\textbf{Strong overall performance does not imply consistent fine-grained capabilities.}
Figure~\ref{fig:subtask_headmap} reveals variation across the 16 subtasks. 
Even the strongest models exhibit weaknesses on individual tasks. 
Qwen3.8-Max, for example, exceeds 70\% accuracy on several subtasks but reaches only 38.0\% on Persistent Counting, while Kimi-K3 drops to 10.0\% on Rearrangement Planning despite strong performance elsewhere. 
Models with similar overall accuracy may still differ across tasks.
Qwen3.5-2B and Ovis2.5-9B achieve comparable overall scores, yet differ considerably across Progress Tracking, Spatial Relations, Constraint Sequencing, and Capacity Planning. 
Overall, strong or similar aggregate performance can still coexist with task-specific weaknesses and capability differences.

\begin{figure*}[t]
    \centering

    \begin{minipage}[t]{0.495\textwidth}
        \centering
        \includegraphics[width=\linewidth]{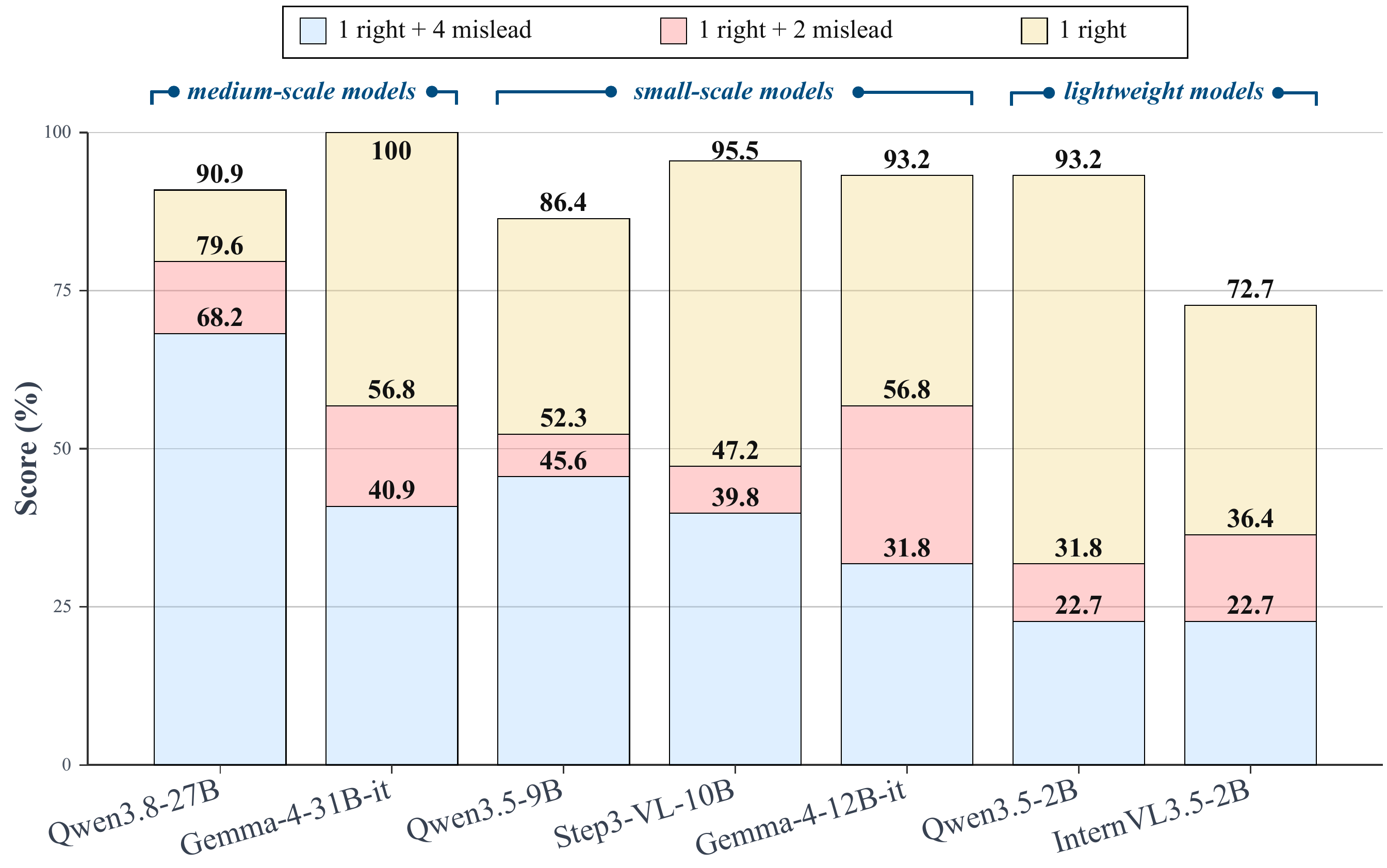}
        \captionof{figure}{Robustness to Increasing Misleading Views across Model Scales.}
        \label{fig:mislead_column}
    \end{minipage}
    \hfill
    \begin{minipage}[t]{0.475\textwidth}
        \centering
        \includegraphics[width=\linewidth]{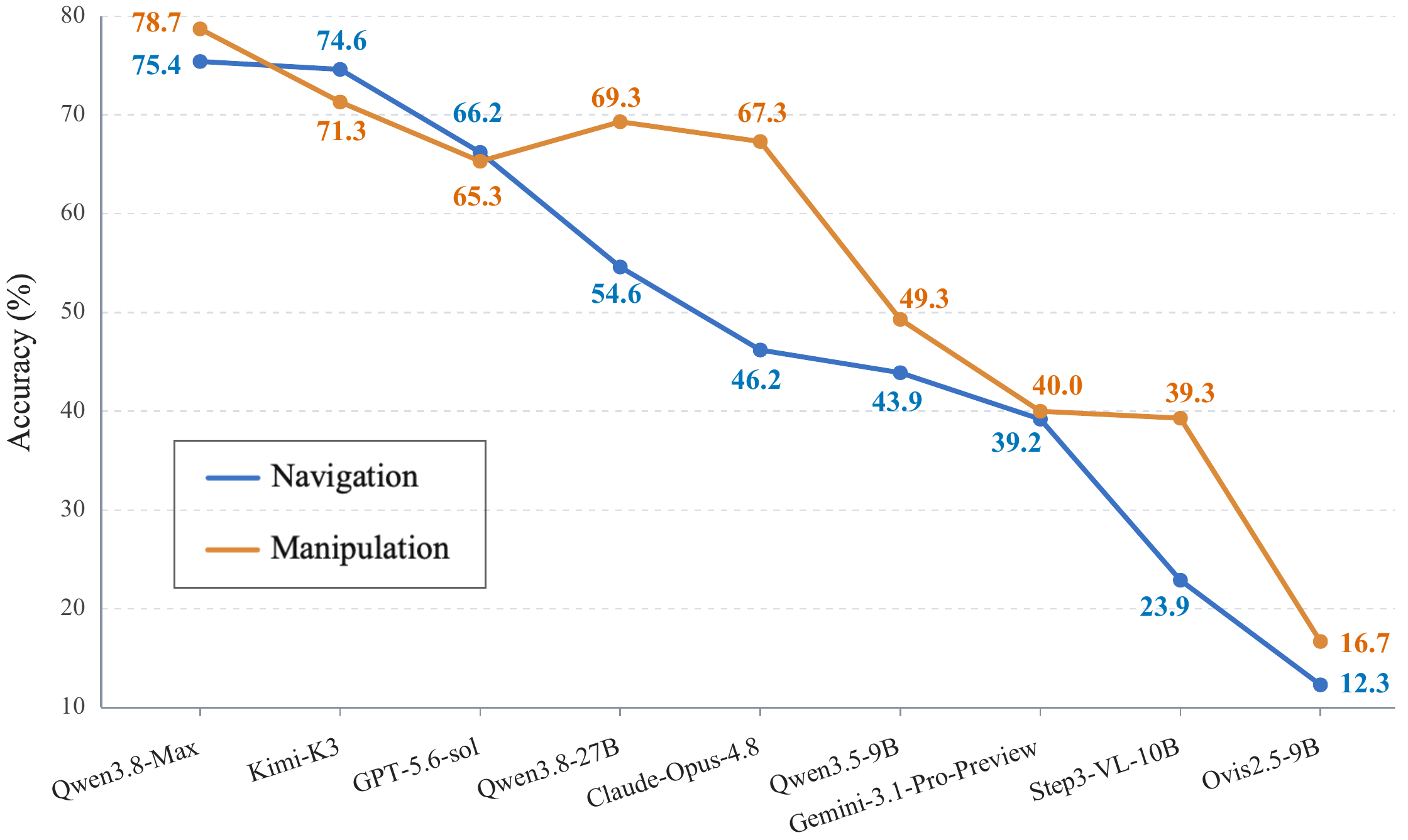}
        \captionof{figure}{Navigation and Manipulation Performance across Different Models.}
        \label{fig:nm_line}
    \end{minipage}

\end{figure*}

\begin{figure*}[t]
    \centering
    \includegraphics[width=\textwidth]{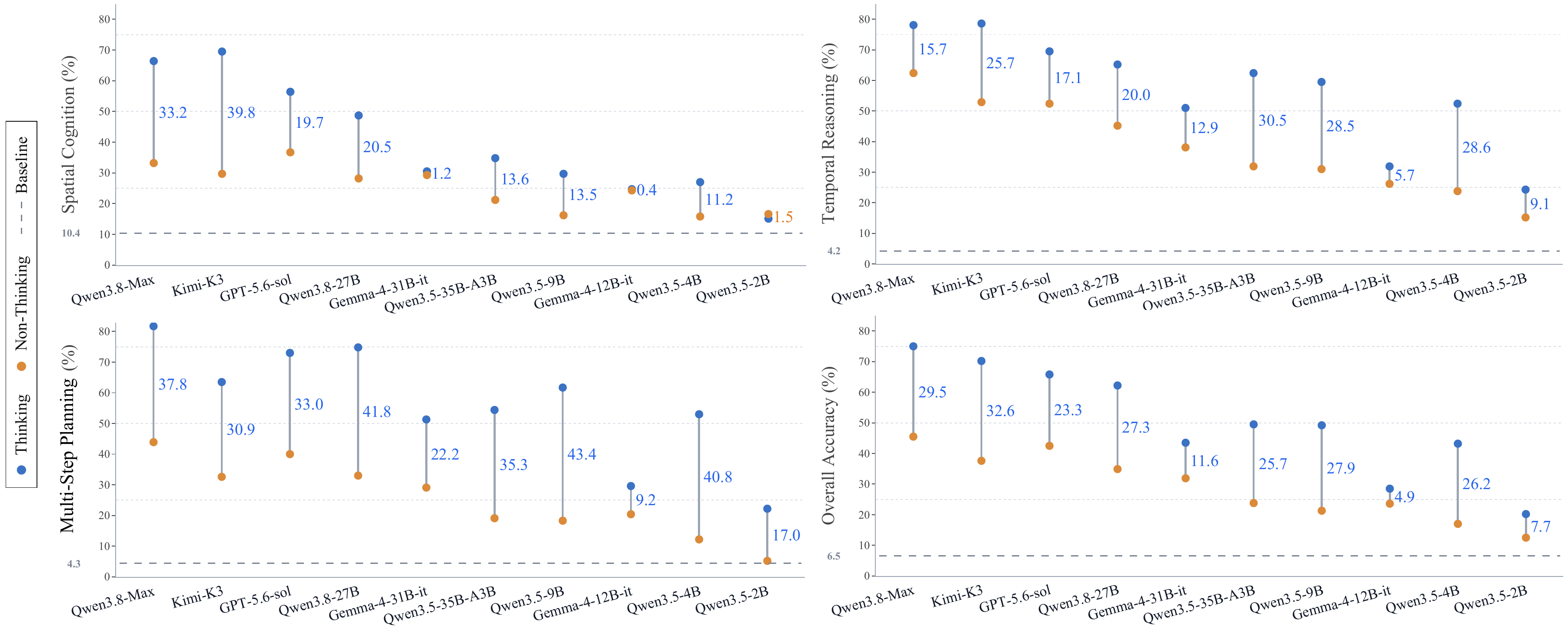}
    \caption{Comparison between Thinking and Non-Thinking modes on Plan-Bench across Spatial Cognition, Temporal Reasoning, Multi-Step Planning, and overall accuracy. Blue and orange markers denote Thinking and Non-Thinking results, while the annotated values indicate their gaps.}
    \label{fig:thinking_study}
\end{figure*}

\textbf{More views do not necessarily lead to better spatial cognition.}
Current VLMs remain susceptible to interference from misleading viewpoints.
We reorganize \textit{Spatial Relations} under the \textit{Spatial Cognition} dimension into three settings: 1 right, 1 right + 2 mislead, and 1 right + 4 mislead, where additional views contain misleading cues.
The results in Figure \ref{fig:mislead_column} show that all seven models perform well when only the informative view is provided, but their accuracy consistently declines as more misleading views are introduced.
Larger models generally show greater robustness to such distractions.
Qwen3.8-27B drops from 90.9\% to 68.2\%, while Qwen3.5-2B falls from 93.2\% to 22.7\%.
Yet scale alone does not explain this robustness, as Gemma-4-31B-it is less robust than the smaller Qwen3.5-9B, suggesting that model architecture and training strategy may also play important roles.

\textbf{Thinking improves performance, with the largest gains in multi-step planning.}
We compare ten models that support both Thinking and Non-Thinking modes on Plan-Bench.
As shown in Figure~\ref{fig:thinking_study}, Thinking improves nearly all dimension-level and overall results, with the only exception of Qwen3.5-2B, which drops by 1.5 percentage points on Spatial Cognition.
Among the three dimensions, Thinking brings the largest gains on Multi-Step Planning, with an average improvement of 31.1 percentage points, compared with 19.4 on Temporal Reasoning and 15.2 on Spatial Cognition.
At the overall level, all models benefit from Thinking, with improvements ranging from 4.9 percentage points on Gemma-4-12B-it to 32.6 points on Kimi-K3.
Despite longer inference time, Thinking enables models to better handle reasoning and planning tasks in embodied environments.

\textbf{Navigation and Manipulation are broadly aligned, with model-specific mismatches.}
We group Plan-Bench into Navigation-related (\textit{Route Planning, Instruction Alignment, and Target Discovery}) and Manipulation-related (\textit{Dependency Planning, Constraint Sequencing, and Occlusion Order}).
As shown in Figure \ref{fig:nm_line}, most models exhibit similar trends.
For example, Qwen3.8-Max, Kimi-K3, and GPT-5.6-sol show balanced performance on the two task groups.
However, some models exhibit biases, with Qwen3.8-27B achieving about 69.3\% on Manipulation task but 54.6\% on Navigation task, while Step3-VL-10B shows a gap of roughly 15 percentage points.
These results further indicate that larger models have greater potential to serve as high-level decision makers in physical environments, with more stable and consistent performance across diverse embodied tasks.

\begin{figure*}[t]
    \centering
    \includegraphics[width=\textwidth]{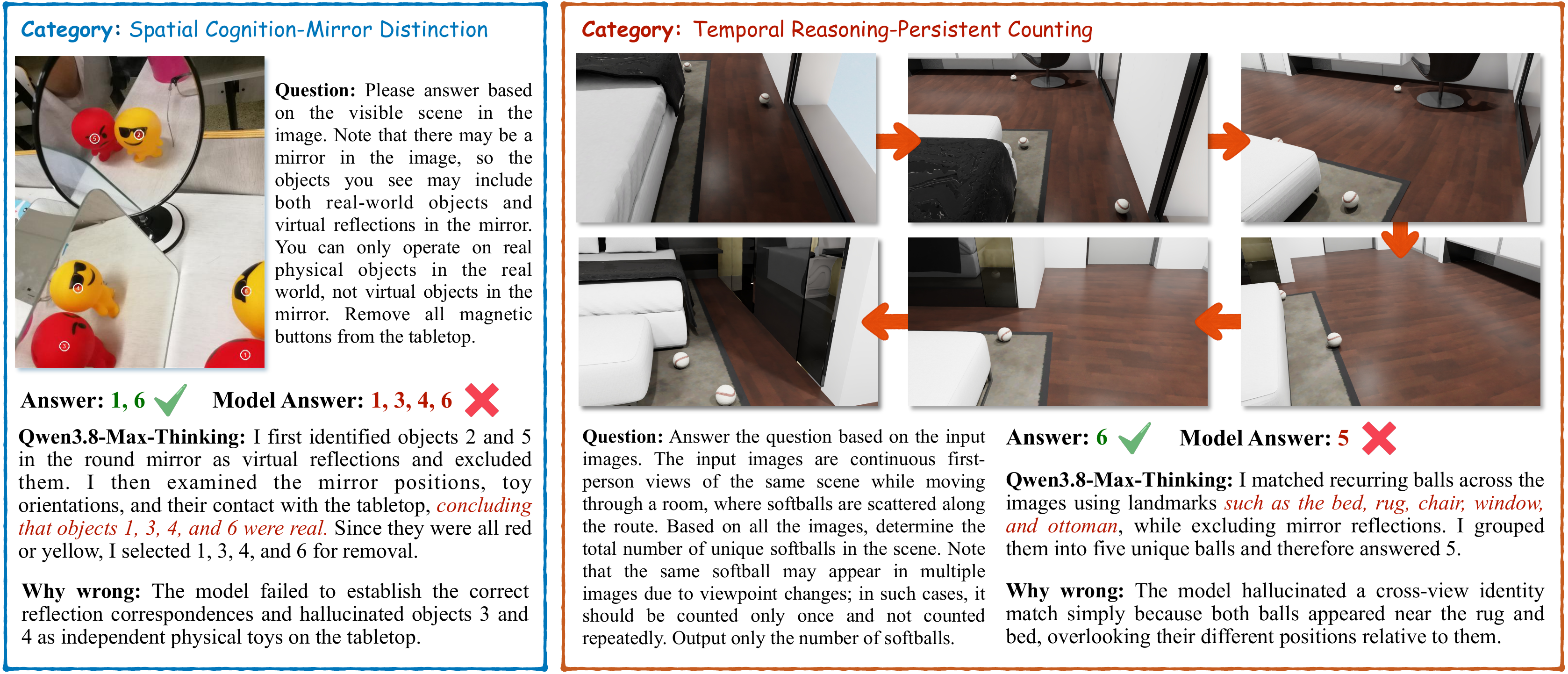}
    \caption{Representative Examples of Hallucination Errors by Qwen3.8-Max}
    \label{fig:hallucination_analysis}
\end{figure*}

\textbf{Hallucination is a source of reasoning and planning errors in the embodied world.}
We analyze Qwen3.8-Max, the best-performing model on Plan-Bench, focusing on failures in \textit{Mirror Distinction, Persistent Counting, and Target Discovery}.
As shown in Figure \ref{fig:hallucination_analysis}, the model fails to distinguish real objects from their reflections in \textit{Mirror Distinction}, hallucinating reflections as physical entities.
In \textit{Persistent Counting}, it incorrectly associates two distinct objects across views due to similar surrounding landmarks, resulting in an incorrect match.
A similar phenomenon also appears in \textit{Target Discovery}.
This suggests that failures arise not only from visual recognition errors, but also from hallucinations during high-level reasoning and multi-step planning.
Reducing such hallucinations may be important for improving the reliability of VLMs in embodied reasoning and planning.

\section{Limitations and Discussion}

We construct Plan-Bench to evaluate the reasoning and planning capabilities of VLMs in embodied tasks across Spatial Cognition, Temporal Reasoning, and Multi-step Planning, while several limitations remain to be further explored and addressed.
\textit{(i) Continually evolving frontier models.}
Although we have evaluated diverse models, frontier systems evolve rapidly and performance may change across versions. 
Future work will track newly released models to assess progress in these capabilities.
\textit{(ii) Limited data scale and task coverage.}
Plan-Bench includes 699 samples across 16 subcategories, while there is still room to expand both the dataset scale and task diversity.
Further expansion will enable a more comprehensive evaluation of reasoning and planning.
\textit{(iii) Potential errors in LLM-as-judge.}
Our LLM-as-judge achieves 100\% agreement with human evaluation on 200 sampled cases, but some errors may still exist in the remaining samples.
We will further explore more robust and programmatic evaluation procedures to improve evaluation accuracy and reliability.

\section{Conclusion}

We introduce Plan-Bench, a lightweight and systematic benchmark for evaluating the high-level reasoning and multi-step planning capabilities of VLMs in embodied scenarios independently of action execution.
Plan-Bench covers Spatial Cognition, Temporal Reasoning, and Multi-Step Planning, spanning 16 subclasses and 699 curated samples.
Evaluation of 28 VLMs shows that current models still have room for improvement, with the best-performing model achieving only 75.0\% accuracy.
Ablation studies further reveal that strong overall performance does not guarantee consistent capabilities across subtasks, and that models remain sensitive to misleading evidence and prone to hallucinations during reasoning.
We also find that scaling generally improves results, while architecture and training may also influence performance, with Thinking bringing additional gains, particularly for Multi-Step Planning.
Our results suggest that larger VLMs show greater potential as high-level planners, but reliable reasoning and planning across diverse embodied tasks remain challenging.
Overall, Plan-Bench provides a principled and easy-to-deploy testbed for tracking progress in these capabilities and assessing VLMs as high-level decision makers for embodied AI.

\bibliography{iclr2027_conference}
\bibliographystyle{iclr2027_conference}

\clearpage
\appendix
\section{Benchmark Details}

\subsection{Task Definitions}

Plan-Bench consists of 16 subtasks that instantiate different reasoning and planning challenges in embodied scenarios.
Below, we provide detailed definitions of each subtask and clarify the specific capability evaluated by each task.

\textbf{Occlusion Order (OO).}
Given a scene containing partially occluded objects, the model is required to infer their relative ordering from visual occlusion cues.
This task evaluates whether the model can reason about object ordering when complete object boundaries are not directly observable.

\textbf{Mirror Distinction (MD).}
The model must distinguish real objects from their reflections and correctly reason about the scene content.
The task is designed to evaluate whether the model can distinguish real objects from reflections and correctly handle reflection-induced visual ambiguity.

\textbf{Action Recovery (AR).}
Given partial information from one viewpoint, the model is asked to recover the corresponding information that is missing or unclear from another viewpoint.
This task examines whether the model can compare observations across multiple views and integrate complementary visual cues to reconstruct the complete scene information.

\textbf{Target Discovery (TD).}
The model must reason about the target’s spatial relationship under mirror reflections and plan a feasible path toward it.
This task evaluates spatial reasoning and path planning in reflection-ambiguous scenes.

\textbf{State Recognition (SR).}
Given multiple visual observations of a scene, the model must integrate information while identifying redundantly observed content.
This task evaluates multi-view scene understanding and the ability to distinguish new evidence from repeated information.

\textbf{Spatial Relations (SpR).}
The model must infer spatial relationships between specified entities from one or multiple views.
The task evaluates spatial reasoning and the ability to integrate relevant evidence while ignoring misleading observations.

\textbf{Progress Tracking (PT).}
Given a sequence depicting an ongoing activity, the model is required to determine the current progress or stage of the activity.
This task evaluates whether the model can track how an embodied procedure evolves over time.

\textbf{Temporal Ordering (TO).}
The model must determine the correct temporal order of events, actions, or scene states from sequential visual evidence.
The task evaluates the ability to recover temporal dependencies and distinguish earlier events from later ones.

\textbf{Persistent Counting (PC).}
The model is asked to count target entities across a visual sequence while maintaining their identity over time.
This task evaluates whether the model can accumulate information across frames without repeatedly counting the same persistent entity.

\textbf{Instruction Alignment (IA).}
Given an instruction together with a sequence of visual observations, the model must identify the observation that corresponds to the specified instruction.
The task evaluates the alignment between language-described actions and their visual realization over time.

\textbf{Motion Retrieval (MR).}
The model is required to identify the motion or action segment that matches a specified query from a temporal sequence.
This task evaluates whether the model can recognize and retrieve task-relevant motion patterns among multiple observed actions.

\textbf{Constraint Sequencing (CS).}
Given a set of actions together with ordering constraints, the model must construct a valid action sequence satisfying all specified constraints.
The task evaluates multi-step planning under explicit requirements on the relative order of actions.

\textbf{Capacity Planning (CP).}
The model is required to produce a feasible plan under capacity limitations on containers or other task-relevant entities.
This task evaluates whether the model can jointly reason about action ordering and limited capacity when constructing a multi-step solution.

\textbf{Rearrangement Planning (RP).}
Given an initial and a target configuration, the model must determine a sequence of actions that transforms the scene into the desired arrangement.
The task evaluates planning capabilities and the consequences of successive rearrangement operations.

\begin{figure*}[t]
    \centering
    \includegraphics[width=\textwidth]{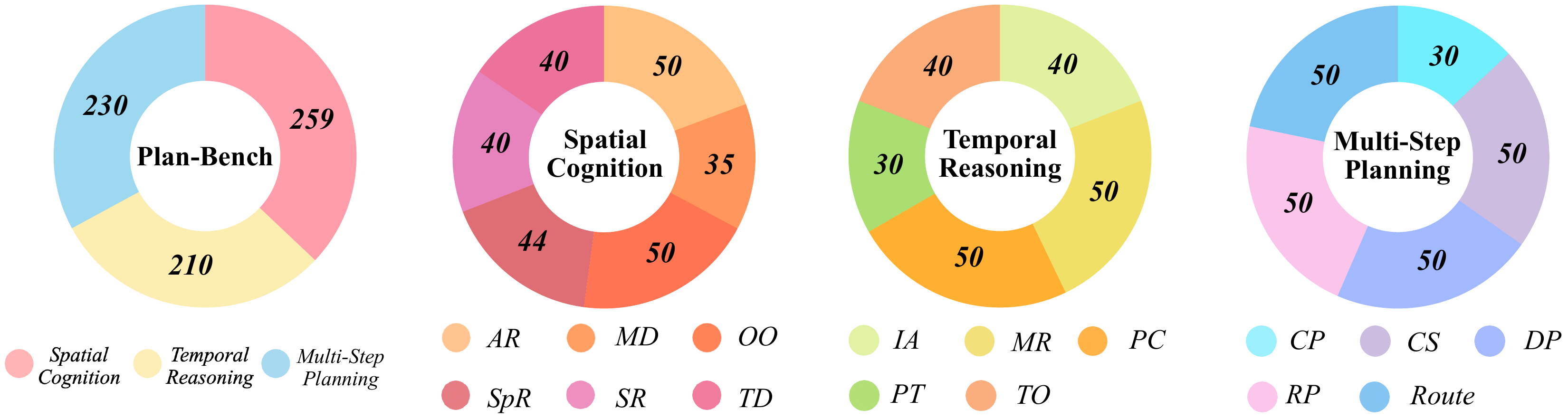}
    \caption{Dataset composition of Plan-Bench across three dimensions and 16 subtasks.}
    \label{fig:subcategory}
\end{figure*}

\begin{table*}[t]
\centering
\caption{
Data sources and question-type distribution across the 16 subtasks in Plan-Bench.
}
\label{tab:data_source}

\begin{small}
\setlength{\tabcolsep}{3.5pt}
\renewcommand{\arraystretch}{1.08}

\begin{tabular}{@{}
>{\centering\arraybackslash}p{0.16\textwidth}
>{\raggedright\arraybackslash}p{0.23\textwidth}
>{\centering\arraybackslash}p{0.08\textwidth}
>{\centering\arraybackslash}p{0.08\textwidth}
>{\raggedright\arraybackslash}p{0.31\textwidth}
@{}}
\toprule

\textbf{Task}
& \textbf{Subtask}
& \textbf{Single}
& \textbf{Open}
& \textbf{Data Source} \\

\midrule

\multirow{6}{*}{\makecell[c]{\textbf{Spatial}\\\textbf{Cognition}}}
& Action Recovery
& 30 & 20 & Robomme~\citep{dai2026robomme} \\ 

& Mirror Distinction
& 0 & 35 &
\makecell[l]{
MSD~\citep{yang2019my}\\
PMD~\citep{lin2020progressive}
} \\

& Occlusion Order
& 18 & 32 & three.js~\citep{danchilla2012three} \\

& Spatial Relations
& 0 & 44 & three.js~\citep{danchilla2012three} \\

& State Recognition
& 20 & 20 & three.js~\citep{danchilla2012three} \\

& Target Discovery
& 40 & 0 & PMD~\citep{lin2020progressive}\\

\midrule

\multirow{5}{*}{\makecell[c]{\textbf{Temporal}\\\textbf{Reasoning}}}
& Instruction Alignment
& 15 & 25 & RxR~\citep{ku2020room}\\

& Motion Retrieval
& 0 & 50 & mevis~\citep{ding2023mevis}\\

& Persistent Counting
& 0 & 50 & OmniGibson~\citep{li2023behavior}\\

& Progress Tracking
& 0 & 30 & captaincook4d~\citep{peddi2024captaincook4d}\\

& Temporal Ordering
& 20 & 20 & captaincook4d~\citep{peddi2024captaincook4d}\\

\midrule

\multirow{5}{*}{\makecell[c]{\textbf{Multi-Step}\\\textbf{Planning}}}
& Capacity Planning
& 30 & 0 & three.js~\citep{danchilla2012three}\\

& Constraint Sequencing
& 0 & 50 & CLEVR~\citep{johnson2017clevr}\\

& Dependency Planning
& 0 & 50 &
\makecell[l]{
Website\\
SpatialQA~\citep{xie2026spatialqa}
} \\

& Rearrangement Planning
& 0 & 50 & OmniGibson~\citep{li2023behavior}\\

& Route Planning
& 10 & 40 & OmniGibson~\citep{li2023behavior}\\

\bottomrule

\end{tabular}
\end{small}

\end{table*}

\begin{table}[t]
\centering
\small

\setlength{\tabcolsep}{2.5pt}
\renewcommand{\arraystretch}{1.02}

\begin{minipage}[t]{0.47\linewidth}
\centering

\caption{Distribution of the number of images per sample.}
\label{tab:image_num}

\begin{tabular}{@{}
>{\raggedright\arraybackslash}p{0.30\linewidth}
>{\centering\arraybackslash}p{0.32\linewidth}
>{\centering\arraybackslash}p{0.25\linewidth}
@{}}
\toprule

\textbf{Images}
& \textbf{Samples}
& \textbf{Ratio}
\\

\midrule

1      & 245 & 35.1\% \\
2--5   & 124 & 17.7\% \\
6--7   & 181 & 25.9\% \\
8--16  & 149 & 21.3\% \\

\midrule

\textbf{Total}
& \textbf{699}
& \textbf{100\%}
\\

\bottomrule
\end{tabular}

\end{minipage}
\hfill
\begin{minipage}[t]{0.47\linewidth}
\centering

\caption{Distribution of image resolutions.}
\label{tab:image_resolution}

\begin{tabular}{@{}
>{\raggedright\arraybackslash}p{0.32\linewidth}
>{\centering\arraybackslash}p{0.30\linewidth}
>{\centering\arraybackslash}p{0.25\linewidth}
@{}}
\toprule

\textbf{Resolution}
& \textbf{Images}
& \textbf{Ratio}
\\

\midrule

$\leq$0.5 MP & 1,430 & 47.6\% \\
0.5--1 MP    & 502   & 16.7\% \\
1--2 MP      & 363   & 12.1\% \\
2--4 MP      & 643   & 21.4\% \\
$\geq$4 MP   & 69    & 2.3\% \\

\midrule

\textbf{Total}
& \textbf{3,007}
& \textbf{100\%}
\\

\bottomrule
\end{tabular}

\end{minipage}

\end{table}

\textbf{Dependency Planning (DP).}
The model must construct an action plan while respecting prerequisite relations among actions or subgoals.
This task evaluates whether the model can identify dependencies and ensure that required preconditions are satisfied before executing dependent actions.

\textbf{Route Planning (Route).}
Given spatial information about an environment, the model must determine a feasible route from the current location to a specified destination.
The task evaluates multi-step navigation planning under spatial connectivity, obstacle, and path-feasibility constraints.

\subsection{Dataset Statistics}       

\textbf{Plan-Bench maintains a balanced distribution across dimensions and subtasks.}
As shown in Figure~\ref{fig:subcategory}, Plan-Bench contains 699 samples distributed across three dimensions, with 259 samples for Spatial Cognition, 210 for Temporal Reasoning, and 230 for Multi-Step Planning.
The samples are further distributed across 16 subtasks, with most subtasks containing between 30 and 50 instances.
Overall, the benchmark maintains a relatively balanced distribution across both dimensions and subtasks, avoiding excessive concentration on any single capability.

\textbf{Plan-Bench exhibits diversity in visual inputs.}
As shown in Tables~\ref{tab:image_num} and~\ref{tab:image_resolution}, the benchmark covers samples containing from 1 to 16 images, with 64.9\% of the samples involving multiple images and 47.2\% containing six or more images.
The 3,007 images also span a wide range of resolutions, from below 0.5 MP to above 4 MP, with 35.8\% exceeding 1 MP.
These statistics highlight the diversity of Plan-Bench in both visual context length and image resolution, rather than relying on a fixed input configuration.

To further characterize the planning complexity of the Multi-Step Planning subset, 
we analyze the number of steps required by the reference plans.
As shown in Table~\ref{tab:planning_horizon_distribution}, the 230 samples span planning 
horizons ranging from 1 to 10 steps.
Among them, 70 samples (30.43\%) require 1--3 steps, while the majority of the samples, 
131 (56.96\%), involve 4--6 planning steps.
In addition, 29 samples (12.61\%) require relatively longer plans of 7--10 steps.
This distribution indicates that Plan-Bench covers planning problems with varying horizon 
lengths, while placing substantial emphasis on multi-step tasks that require models to 
reason over several interdependent actions rather than making only short-horizon decisions.

\begin{table}[t]
\centering
\caption{Distribution of planning lengths in the Multi-Step Planning subset of Plan-Bench.}
\label{tab:planning_horizon_distribution}
\setlength{\tabcolsep}{14pt}
\renewcommand{\arraystretch}{0.9}
\begin{tabular}{lcc}
\toprule
\textbf{Steps} & \textbf{Samples} & \textbf{Percentage} \\
\midrule
1--3 steps  & 70  & 30.43\% \\
4--6 steps  & 131 & 56.96\% \\
7--10 steps & 29  & 12.61\% \\
\midrule
\textbf{Total} & \textbf{230} & \textbf{100\%} \\
\bottomrule
\end{tabular}
\end{table}

\begin{table}[t]
\centering
\caption{Category distribution before and after sample filtering.}
\label{tab:filtering_task_distribution}
\setlength{\tabcolsep}{12pt}
\renewcommand{\arraystretch}{0.9}
\begin{tabular}{lccc}
\toprule
\textbf{Category} & 
\textbf{Before Filtering} & 
\textbf{Removed} &
\textbf{Retained} \\
\midrule

Spatial Cognition 
& 952 & 693 & 259 \\

Temporal Reasoning 
& 768 & 558 & 210 \\

Multi-Step Planning 
& 860 & 630 & 230 \\

\midrule
\textbf{Total} 
& \textbf{2580} & \textbf{1881} & \textbf{699} \\

\bottomrule
\end{tabular}
\end{table}

\begin{table}[t]
\centering
\caption{Statistics of sample filtering during Plan-Bench construction.}
\label{tab:filtering_statistics}
\setlength{\tabcolsep}{12pt}
\renewcommand{\arraystretch}{1.0}
\begin{tabular}{lcc}
\toprule
\textbf{Filtering Reason} & \textbf{Removed Samples} & \textbf{Percentage} \\
\midrule
Insufficient visual evidence or ambiguous answers
& 637 & 33.9\% \\
Incorrect annotation or reference answer 
& 364 & 19.4\% \\
Capability mismatch with target task 
& 297 & 15.8\% \\
Duplicate or highly correlated scenes
& 512 & 27.2\% \\
Other quality issues 
& 71 & 3.8\% \\

\midrule
\textbf{Total removed} 
& \textbf{1881} & \textbf{72.91\%} \\
\textbf{Final retained} 
& \textbf{699} & \textbf{27.09\%} \\

\bottomrule
\end{tabular}
\end{table}

\subsection{Dataset Filtering Procedure}
\label{app:dataset_filtering}

We perform a multi-stage filtering procedure to improve the quality and
reliability of Plan-Bench. Starting from 2,580 candidate QA pairs, we remove
samples that fail one or more quality requirements and finally obtain 699
validated samples. The entire filtering process is conducted before model
evaluation and is independent of model performance.

Each candidate sample is examined according to five quality considerations:
(1) visual sufficiency, where the answer must be determinable from the provided
visual evidence; (2) annotation correctness, including the validity of reference
answers and answer choices; (3) capability alignment, ensuring that the sample
measures the intended reasoning or planning ability; (4) redundancy control,
which removes highly correlated samples originating from similar scenes, videos,
or viewpoints; and (5) other quality issues identified during manual review.
The detailed statistics of removed samples are reported in
Table~\ref{tab:filtering_statistics}.

For samples collected from videos or multi-view environments, redundancy
filtering is performed at the scene level rather than only at the QA level.
Samples sharing the same underlying scene, highly similar viewpoints, or
overlapping temporal segments are grouped together. When multiple samples
evaluate equivalent reasoning patterns, only the highest-quality instance is
retained to avoid over-representing specific environments.

To examine whether filtering introduces unintended changes in benchmark
composition, we compare the category distribution before and after filtering.
As shown in Table~\ref{tab:filtering_task_distribution}, the retention ratios
remain similar across Spatial Cognition, Temporal Reasoning, and Multi-Step
Planning. The final benchmark maintains coverage across all three dimensions
and 16 subtasks while removing low-quality or redundant samples.

\subsection{Data Sources}

\textbf{Plan-Bench combines diverse data sources and question formats across subtasks.}
As shown in Table~\ref{tab:data_source}, the 16 subtasks are constructed from a mixture of existing datasets, controlled generation tools, simulation environments, and web-sourced data.
Different sources are selected according to the capability targeted by each subtask, ranging from temporal activities to spatial reasoning and multi-step planning.
The benchmark also includes single-choice, open-ended, and mixed-format subtasks, providing diverse evaluation settings rather than relying on a uniform question format.

\clearpage
\section{Evaluation Protocol}

\subsection{Model Prompt Templates}
\label{app:base_prompt}

Since Plan-Bench contains both single-choice and open-ended questions, we implement a unified prompting function to support both question types within the same evaluation pipeline.
As shown below, all samples share a common \textit{Base Prompt}, which specifies the general response behavior and output requirements.
We then append a question-type-specific instruction according to the sample format: the \textit{Single-Choice Prompt} requires the model to return only one option letter, while the \textit{Open-Ended Prompt} requires only the requested structured answer.
This design provides a consistent prompting framework while preserving the output requirements of different question types.

\begin{promptbox}{Base Prompt}

Answer directly. Do not explain your reasoning.
For single-choice questions, answer with one option letter only, such as A.
For open questions, follow the required answer format exactly.

\end{promptbox}

\begin{promptbox}{Single-Choice Prompt}

This is a single-choice question. Return one option letter only.

\end{promptbox}

\begin{promptbox}{Open-Ended Prompt}

This is an open question. Return only the requested structured answer.

\end{promptbox}

\subsection{LLM-as-Judge Prompt}
\label{app:judge_prompt}

We use DeepSeek-V4-Flash as an LLM-based judge to evaluate model responses.
For each sample, the judge receives the model output, ground-truth answer, and an optional answer hint specifying additional valid alternatives or equivalence conditions.
A response is labeled \texttt{True} when it is semantically equivalent to the ground-truth answer or satisfies an equivalence condition explicitly defined by the answer hint; otherwise, it is labeled \texttt{False}.
The judge is instructed to evaluate only answer correctness and to return no additional explanation.

\begin{promptbox}{System Prompt (Answer Equivalence Judge)}

You are a precise answer-equivalence judge for evaluating model responses.

Your task is to determine whether a model output should be considered correct with respect to the ground-truth answer and the provided answer hint.

The \textbf{ground-truth answer} specifies the reference answer, while the \textbf{answer hint} provides additional valid alternatives or equivalence conditions that should be accepted.

Judge the model output according to the following rules:

\begin{enumerate}[
    leftmargin=1.6em,
    labelsep=0.45em,
    itemsep=4pt,
    topsep=3pt,
    parsep=0pt
]

\item \textbf{Semantic Equivalence:}
Return \texttt{True} if the model output expresses the same answer
as the ground-truth answer, even if the wording or formatting differs.

\item \textbf{Answer-Hint Equivalence:}
Return \texttt{True} if the model output satisfies any valid
alternative or equivalence condition explicitly specified in the
answer hint, including acceptable order swaps, position exchanges,
alternative object names, or equivalent action descriptions.

\item \textbf{Partial Correctness:}
Return \texttt{False} if the response is only partially correct,
omits required information, or contains an incorrect component.

\item \textbf{Contradictory or Additional Content:}
Return \texttt{False} if the response contains information that
contradicts the ground-truth answer or violates the conditions
specified by the answer hint.

\item \textbf{Formatting Differences:}
Ignore superficial differences in capitalization, punctuation,
spacing, or other formatting when they do not change the meaning
of the answer.

\end{enumerate}

Do not introduce alternative interpretations beyond those supported by the ground-truth answer and answer hint.
Do not infer missing information.
Do not evaluate the reasoning process; evaluate only the final answer.

Return only one of the following:

\texttt{True}

or

\texttt{False}

Do not provide any explanation or additional text.

\end{promptbox}

\begin{promptbox}{User Prompt (Answer Verification)}

Determine whether the model output is equivalent to the ground-truth answer under the conditions specified by the answer hint.

\textbf{Ground-truth answer:}

\texttt{\{ground\_truth\}}

\textbf{Answer hint:}

\texttt{\{answer\_hint\}}

\textbf{Model output:}

\texttt{\{model\_output\}}

Return only \texttt{True} or \texttt{False}.

\end{promptbox}

\subsection{Answer Hint and Equivalence Rules}

\begin{figure}[t]
    \centering
    \includegraphics[width=\linewidth]{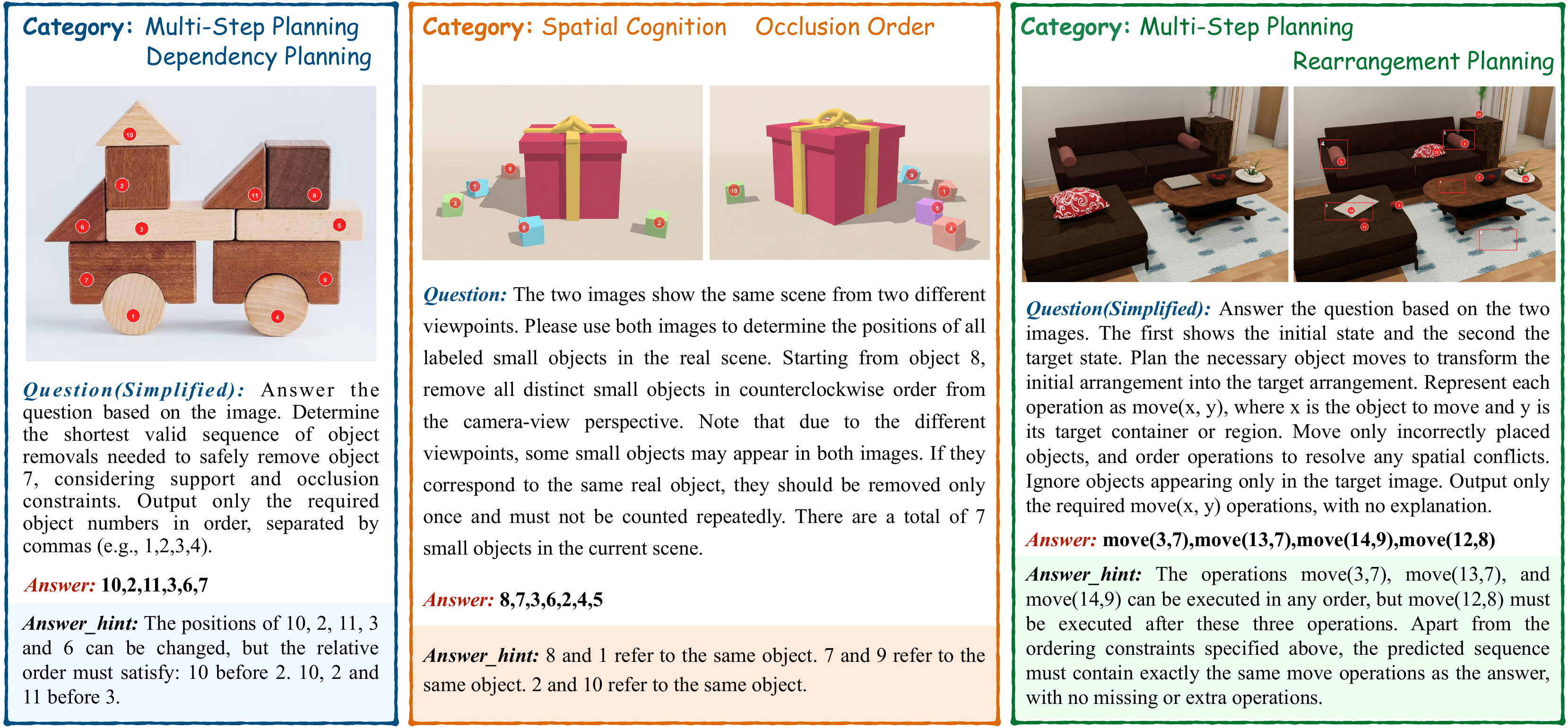}
    \caption{Examples of answer hints for multiple valid answers.}
    \label{fig:answer_hint}
\end{figure}

\textbf{Answer hints handle multiple valid answers in open-ended questions.}
Among the open-ended questions in Plan-Bench, 486 samples require answer hints across 13 subcategories, including \textit{Occlusion Order}, \textit{Mirror Distinction}, \textit{State Recognition}, \textit{Rearrangement Planning}, and \textit{Dependency Planning}.
These hints specify valid equivalence conditions beyond the reference answer, such as interchangeable object references, flexible orderings, position exchanges, and equivalent action sequences.
As illustrated in Figure~\ref{fig:answer_hint}, the answer hint is provided only to the LLM judge, allowing semantically equivalent predictions to be accepted without exposing additional information to the evaluated model.

\subsection{Human Verification of the Judge}
To further assess the reliability of our LLM-based evaluation, we conduct a human verification study on 200 responses generated by Qwen3.8-Max, Gemma-4-31B-it, and Step3-VL-10B.
We adopt stratified sampling to ensure diverse and balanced coverage across the three evaluation dimensions and their corresponding subclasses, selecting 67 samples from \textit{Spatial Cognition}, 67 from \textit{Temporal Reasoning}, and 66 from \textit{Multi-Step Planning}.
Among the 200 sampled instances, 133 contain answer hints specifying acceptable alternatives or equivalence conditions, such as alternative object names, interchangeable orderings, or equivalent action sequences, while the remaining 67 have no answer hints.
All 200 responses are independently evaluated by human annotators and DeepSeek-V4-Flash based on the same ground-truth answers and answer hints.
The two evaluations agree on all 200 samples, achieving 100\% agreement, including the 133 cases involving answer hints.
These results demonstrate that our LLM-based judge can reliably account for equivalent valid answers and provide consistent evaluation for Plan-Bench.

\subsection{Model Configurations}

We evaluate open-weight VLMs through VLMEvalKit using their corresponding model wrappers and checkpoints. Unless specified below, we retain the decoding and image-processing settings defined by each wrapper. Open-weight models are evaluated on NVIDIA A800 GPUs in BF16 precision. All models receive the prompt templates described in Appendix~\ref{app:base_prompt}, while proprietary models are evaluated through API interfaces under the same question and answer protocol.

\textbf{Qwen series.}
Qwen3-VL-2B-Instruct, Qwen3-VL-8B-Instruct, and Qwen3-VL-30B-A3B-Instruct use the Qwen3VLChat wrapper with custom prompts disabled. Their configurations set temperature to 0.7, top-$p$ to 0.8, top-$k$ to 20, presence penalty to 1.5, and the maximum number of new tokens to 16,384. We use the same wrapper for Qwen3.5-2B, 4B, 9B, 27B, and 35B-A3B, as well as Qwen3.8-27B, with temperature 1.0, top-$p$ 0.95, top-$k$ 20, and presence penalty 1.5. For these Qwen3.5 and Qwen3.8 models, the initial generation limit is 16,384 tokens. After the initial generation, we instruct each model to summarize its generated response and present the final answer within an additional 512-token budget. This additional budget is used only for answer condensation, extraction, and formatting based on the already generated content, rather than for further substantive reasoning or revising the underlying answer. Where supported, Thinking and Non-Thinking evaluations use the same checkpoint with the corresponding mode selected in the wrapper.

\textbf{Gemma-4 series.}
Gemma-4-12B-it and Gemma-4-31B-it use the Gemma4 wrapper with the corresponding model processor and chat template. The wrapper uses greedy decoding with a maximum of 16,384 new tokens. For Thinking and Non-Thinking evaluations, we set the chat template's \texttt{enable\_thinking} option accordingly and extract the final answer from the generated response.

\textbf{InternVL3.5 series.}
InternVL3.5-2B, InternVL3.5-8B, and InternVL3.5-30B-A3B use the InternVLChat V2.0 wrapper with the vLLM backend enabled. We retain the wrapper's default image processing and generation settings for these models.

\textbf{Step3-VL series.}
Step3-VL-10B uses a local vLLM inference wrapper with greedy decoding, a maximum of 16,384 new tokens, and support for up to 24 images per prompt. Its DeepSeek-R1-style parser separates the reasoning trace from the final answer; only the final answer is submitted for evaluation.

\textbf{GLM-4.6V series.}
GLM-4.6V-Flash uses a local vLLM wrapper with Thinking enabled. Its configuration sets temperature to 0.8, top-$p$ to 0.6, top-$k$ to 2, repetition penalty to 1.1, and the maximum number of new tokens to 16,384.

\textbf{Ovis series.}
Ovis2-8B and Ovis2.5-9B are evaluated using their corresponding VLMEvalKit wrappers. We retain their wrapper-defined image preprocessing and generation settings.

\textbf{MiMo-VL series.}
MiMo-VL-7B-SFT uses the Qwen2VLChat wrapper with custom prompts disabled and vLLM inference. Its image-size bounds are set to $1280\times28\times28$ and $16384\times28\times28$ pixels; the remaining generation settings follow the wrapper configuration.

\textbf{NEO series.}
NEO-2B-SFT and NEO-9B-SFT use the NEOChat wrapper with a patch size of 16, a downsampling ratio of 0.5, image-size bounds of $1280\times32\times32$ and $4096\times32\times32$ pixels, and the maximum number of new tokens to 16,384. We retain the wrapper's default generation settings. NEO-ov is evaluated under the same benchmark protocol, although its model-specific configuration is not included in the supplied files.

\textbf{SenseNova series.}
SenseNova-U1 is evaluated under the common Plan-Bench prompting and answer protocol, and the maximum number of new tokens to 16,384. Its model-specific inference settings are not specified in the supplied configuration.

\textbf{Qwen3.8-max.}
Qwen3.8-max is evaluated through an API interface with a maximum response length of 16,384 tokens. The supplied configuration does not specify a temperature override.

\textbf{GPT-5.6-sol.}
GPT-5.6-sol is evaluated through an API interface with a maximum response length of 16,384 tokens. The supplied configuration does not specify a temperature override.

\textbf{Gemini-3.1-Pro-Preview.}
Gemini-3.1-Pro-Preview is evaluated through an API interface with temperature set to 0 and a maximum response length of 16,384 tokens.

\textbf{Claude-Opus-4.8.}
Claude-Opus-4.8 is evaluated through an API interface with temperature set to 0 and a maximum response length of 16,384 tokens. 

\textbf{Kimi-K3.}
Kimi-K3 is evaluated under the common Plan-Bench prompting and answer protocol. Its model-specific API settings are not specified in the supplied configuration files.

\renewcommand{\rrw}[7]{%
  #1 & #2 & #3 & #4 & #5 & #6 & #7 \\
}
\renewcommand{\grp}[1]{%
  \rowcolor{myblue}
  \multicolumn{7}{l}{\textbf{$\blacktriangledown$ #1}} \\
}

\clearpage
\section{Additional Results}
\subsection{Results on All 16 Subtasks}

Tables~\ref{tab:spatial_subtask_results}--\ref{tab:planning_subtask_results} report model accuracy on all 16 subtasks of Plan-Bench, revealing differences that are hidden by the three dimension-level scores.
In Spatial Cognition, GPT-5.6-sol achieves 82.9\% on Mirror Distinction and 87.5\% on State Recognition, but only 30.0\% on Target Discovery and 31.3\% on Spatial Relations.
Gemini-3.1-Pro-Preview shows a contrasting profile, reaching 75.0\% on Spatial Relations but only 22.0\% on Occlusion Order.
Qwen3.5-27B also varies sharply across the category, with 56.8\% on Spatial Relations but 18.0\% on Occlusion Order.
These contrasts indicate that success on one form of spatial evidence does not reliably extend to the other visual reasoning demands within the same dimension, even when the tasks draw on similar visual scenes and viewpoints that expose different but complementary scene cues about object locations.

Temporal Reasoning likewise exhibits substantial variation between subtasks.
Qwen3.5-27B achieves 96.7\% on Progress Tracking and 80.0\% on Temporal Ordering, but only 28.0\% on Persistent Counting.
GPT-5.6-sol follows a similar pattern, scoring 93.3\% on Progress Tracking and 30.0\% on Persistent Counting.
Even Qwen3.5-35B-A3B, which reaches 40.0\% on Persistent Counting, performs much better on Progress Tracking at 86.7\%.
These gaps suggest that inferring task progress and event order is easier for these models than maintaining accurate counts when objects or actions recur across visual sequences and must be distinguished from previous observations or distractors that appear between relevant activity steps in multi-frame observations.

In Multi-Step Planning, strong results on one type of constraint do not consistently transfer to other planning problems.
Gemini-3.1-Pro-Preview scores 93.3\% on Capacity Planning but only 10.0\% on Rearrangement Planning.
Similarly, Qwen3.5-27B scores 90.0\% on Capacity Planning but 44.0\% on Route Planning.
Rearrangement Planning is particularly difficult for several smaller models: InternVL3.5-8B and NEO-9B-SFT both score 0.0\% despite obtaining 30.0\% and 53.3\% on Capacity Planning, respectively.
Thus, the detailed results expose task-specific weaknesses in spatial reasoning, temporal tracking, and multi-step planning that aggregate accuracy cannot capture.

\subsection{Additional Error Analysis}
Figure~\ref{fig:hallucination_2} presents two additional Target Discovery failures by Qwen3.8-Max. In both cases, the target appears in a mirror, so selecting the correct action requires the model to associate the reflection with a real object and infer that object's location. In the first example, the model notices the similarity between the reflected picture and the directly visible picture, but speculates that another identical picture exists elsewhere. It therefore selects an action that leads away from the actual target. In the second example, the model likewise treats the reflected and visible paper towel dispensers as two separate objects, incorrectly concluding that the target is behind the camera.

These examples show how hallucination can arise when the model fails to establish the correspondence between a reflection and a physical object. Although it recognizes their visual similarity, it hallucinates a second, identical object and reasons about its nonexistent location. This mistaken interpretation then leads to an incorrect navigation decision, showing how object hallucination can propagate from scene understanding to action planning in Target Discovery.

\begin{table*}[!p]
\caption{Performance of different VLMs on the six Spatial Cognition subtasks of Plan-Bench.}
\label{tab:spatial_subtask_results}
\centering
\begin{small}
\setlength{\tabcolsep}{2.5pt}
\renewcommand{\arraystretch}{1.08}

\begin{tabular}{@{}
>{\raggedright\arraybackslash}p{0.32\textwidth}|
*{5}{>{\centering\arraybackslash}p{0.08\textwidth}|}
>{\centering\arraybackslash}p{0.08\textwidth}
@{}}
\toprule
\textbf{Model}
& \textbf{OO} & \textbf{MD} & \textbf{AR}
& \textbf{TD} & \textbf{SR} & \textbf{SpR} \\
\midrule

\grp{Commercial Models}
\rrw{GPT-5.6-sol}{58.0}{82.9}{54.0}{30.0}{87.5}{31.3}
\rrw{Claude-Opus-4.8}{64.0}{74.3}{28.0}{27.5}{42.5}{22.7}
\rrw{Gemini-3.1-Pro-Preview}{22.0}{45.7}{28.0}{30.0}{42.5}{75.0}
\midrule

\grp{Large Open-Weight Models}
\rrw{Qwen3.5-27B}{18.0}{54.3}{36.0}{27.5}{47.5}{56.8}
\rrw{Qwen3.5-35B-A3B}{12.0}{51.4}{20.0}{35.0}{50.0}{50.0}
\rrw{Qwen3-VL-30B-A3B}{18.0}{42.9}{24.0}{7.5}{25.0}{20.5}
\rrw{InternVL3.5-30B-A3B}{6.0}{2.9}{24.0}{10.0}{20.0}{22.7}
\midrule

\grp{Medium Open-Weight Models}
\rrw{Qwen3.5-9B}{6.0}{45.7}{18.0}{30.0}{35.0}{52.3}
\rrw{Gemma-4-12B-it}{10.0}{37.1}{24.0}{25.0}{22.5}{34.1}
\rrw{InternVL3.5-8B}{22.0}{14.3}{20.0}{15.0}{15.0}{27.3}
\rrw{NEO-9B-SFT}{18.0}{0.0}{30.0}{20.0}{22.5}{43.2}
\rrw{MiMo-VL-7B-SFT}{14.0}{40.0}{14.0}{15.0}{17.5}{22.7}
\rrw{SenseNova-U1}{8.0}{8.6}{0.0}{20.0}{45.5}{30.0}
\rrw{Qwen3-VL-8B-Instruct}{24.0}{11.4}{8.0}{22.5}{20.0}{20.5}
\rrw{NEO-ov}{24.0}{0.0}{16.0}{12.5}{17.5}{36.4}
\rrw{Ovis2-8B}{8.0}{0.0}{14.0}{20.0}{20.0}{27.3}
\midrule

\grp{Small Open-Weight Models}
\rrw{Qwen3.5-4B}{6.0}{40.0}{20.0}{17.5}{40.0}{45.5}
\rrw{Qwen3-VL-2B-Instruct}{16.0}{2.9}{16.0}{12.5}{15.0}{20.5}
\rrw{InternVL3.5-2B}{18.0}{0.0}{16.0}{15.0}{10.0}{20.5}
\rrw{NEO-2B-SFT}{10.0}{2.9}{10.0}{10.0}{12.5}{15.9}
\bottomrule
\end{tabular}
\end{small}

\vspace{0.25in}

\renewcommand{\rrw}[6]{%
  #1 & #2 & #3 & #4 & #5 & #6 \\
}
\renewcommand{\grp}[1]{%
  \rowcolor{myblue}
  \multicolumn{6}{l}{\textbf{$\blacktriangledown$ #1}} \\
}

\caption{Performance of different VLMs on the five Temporal Reasoning subtasks of Plan-Bench.}
\label{tab:temporal_subtask_results}
\centering
\begin{small}
\setlength{\tabcolsep}{2.5pt}
\renewcommand{\arraystretch}{1.08}

\begin{tabular}{@{}
>{\raggedright\arraybackslash}p{0.32\textwidth}|
*{4}{>{\centering\arraybackslash}p{0.10\textwidth}|}
>{\centering\arraybackslash}p{0.10\textwidth}
@{}}
\toprule
\textbf{Model}
& \textbf{PT} & \textbf{TO} & \textbf{PC}
& \textbf{IA} & \textbf{MR} \\
\midrule

\grp{Commercial Models}
\rrw{GPT-5.6-sol}{93.3}{75.0}{30.0}{80.0}{82.0}
\rrw{Claude-Opus-4.8}{93.3}{50.0}{34.0}{40.0}{44.0}
\rrw{Gemini-3.1-Pro-Preview}{86.7}{67.5}{34.0}{47.5}{62.0}
\midrule

\grp{Large Open-Weight Models}
\rrw{Qwen3.5-27B}{96.7}{80.0}{28.0}{72.5}{68.0}
\rrw{Qwen3.5-35B-A3B}{86.7}{80.0}{40.0}{55.0}{62.0}
\rrw{Qwen3-VL-30B-A3B}{40.0}{10.0}{10.0}{17.5}{36.0}
\rrw{InternVL3.5-30B-A3B}{33.3}{22.5}{10.0}{20.0}{42.0}
\midrule

\grp{Medium Open-Weight Models}
\rrw{Qwen3.5-9B}{93.3}{72.5}{28.0}{62.5}{58.0}
\rrw{Gemma-4-12B-it}{80.0}{7.5}{14.0}{22.5}{48.0}
\rrw{InternVL3.5-8B}{26.7}{25.0}{24.0}{35.0}{30.0}
\rrw{NEO-9B-SFT}{10.0}{15.0}{20.0}{7.5}{36.0}
\rrw{MiMo-VL-7B-SFT}{13.3}{22.5}{12.0}{17.5}{38.0}
\rrw{SenseNova-U1}{30.0}{7.5}{8.0}{7.5}{26.0}
\rrw{Qwen3-VL-8B-Instruct}{20.0}{20.0}{8.0}{15.0}{34.0}
\rrw{NEO-ov}{13.3}{12.5}{14.0}{17.5}{28.0}
\rrw{Ovis2-8B}{30.0}{15.0}{14.0}{12.5}{26.0}
\midrule

\grp{Small Open-Weight Models}
\rrw{Qwen3.5-4B}{86.7}{67.5}{24.0}{40.0}{58.0}
\rrw{Qwen3-VL-2B-Instruct}{13.3}{15.0}{14.0}{10.0}{30.0}
\rrw{InternVL3.5-2B}{10.0}{17.5}{18.0}{15.0}{28.0}
\rrw{NEO-2B-SFT}{10.0}{15.0}{2.0}{2.5}{8.0}
\bottomrule
\end{tabular}
\end{small}
\end{table*}
\clearpage

\renewcommand{\rrw}[6]{%
  #1 & #2 & #3 & #4 & #5 & #6 \\
}
\renewcommand{\grp}[1]{%
  \rowcolor{myblue}
  \multicolumn{6}{l}{\textbf{$\blacktriangledown$ #1}} \\
}

\begin{table*}[!p]
\caption{Performance of different VLMs on the five Multi-Step Planning subtasks of Plan-Bench.}
\label{tab:planning_subtask_results}
\centering
\begin{small}
\setlength{\tabcolsep}{2.5pt}
\renewcommand{\arraystretch}{1.08}

\begin{tabular}{@{}
>{\raggedright\arraybackslash}p{0.32\textwidth}|
*{4}{>{\centering\arraybackslash}p{0.10\textwidth}|}
>{\centering\arraybackslash}p{0.10\textwidth}
@{}}
\toprule
\textbf{Model}
& \textbf{CS} & \textbf{CP} & \textbf{RP}
& \textbf{DP} & \textbf{Route} \\
\midrule

\grp{Commercial Models}
\rrw{GPT-5.6-sol}{52.0}{86.7}{62.0}{86.0}{84.0}
\rrw{Claude-Opus-4.8}{70.0}{90.0}{34.0}{68.0}{66.0}
\rrw{Gemini-3.1-Pro-Preview}{42.0}{93.3}{10.0}{56.0}{40.0}
\midrule

\grp{Large Open-Weight Models}
\rrw{Qwen3.5-27B}{82.0}{90.0}{62.0}{62.0}{44.0}
\rrw{Qwen3.5-35B-A3B}{62.0}{86.7}{54.0}{52.0}{30.0}
\rrw{Qwen3-VL-30B-A3B}{24.0}{33.3}{10.0}{18.0}{0.0}
\rrw{InternVL3.5-30B-A3B}{26.0}{36.7}{0.0}{6.0}{10.0}
\midrule

\grp{Medium Open-Weight Models}
\rrw{Qwen3.5-9B}{72.0}{86.7}{50.0}{70.0}{40.0}
\rrw{Gemma-4-12B-it}{38.0}{26.7}{6.0}{54.0}{22.0}
\rrw{InternVL3.5-8B}{22.0}{30.0}{0.0}{10.0}{6.0}
\rrw{NEO-9B-SFT}{10.0}{53.3}{0.0}{8.0}{8.0}
\rrw{MiMo-VL-7B-SFT}{24.0}{26.7}{2.0}{4.0}{18.0}
\rrw{SenseNova-U1}{24.0}{30.0}{0.0}{36.0}{18.0}
\rrw{Qwen3-VL-8B-Instruct}{26.0}{36.7}{8.0}{8.0}{12.0}
\rrw{NEO-ov}{14.0}{50.0}{0.0}{4.0}{10.0}
\rrw{Ovis2-8B}{16.0}{50.0}{0.0}{4.0}{10.0}
\midrule

\grp{Small Open-Weight Models}
\rrw{Qwen3.5-4B}{72.0}{80.0}{28.0}{56.0}{40.0}
\rrw{Qwen3-VL-2B-Instruct}{10.0}{30.0}{0.0}{2.0}{8.0}
\rrw{InternVL3.5-2B}{8.0}{20.0}{0.0}{2.0}{4.0}
\rrw{NEO-2B-SFT}{2.0}{30.0}{0.0}{8.0}{4.0}
\bottomrule
\end{tabular}
\end{small}

\vspace{0.25in}

\begin{minipage}{\textwidth}
    \centering
    \includegraphics[
        width=0.99\textwidth,
        height=0.28\textheight,
        keepaspectratio
    ]{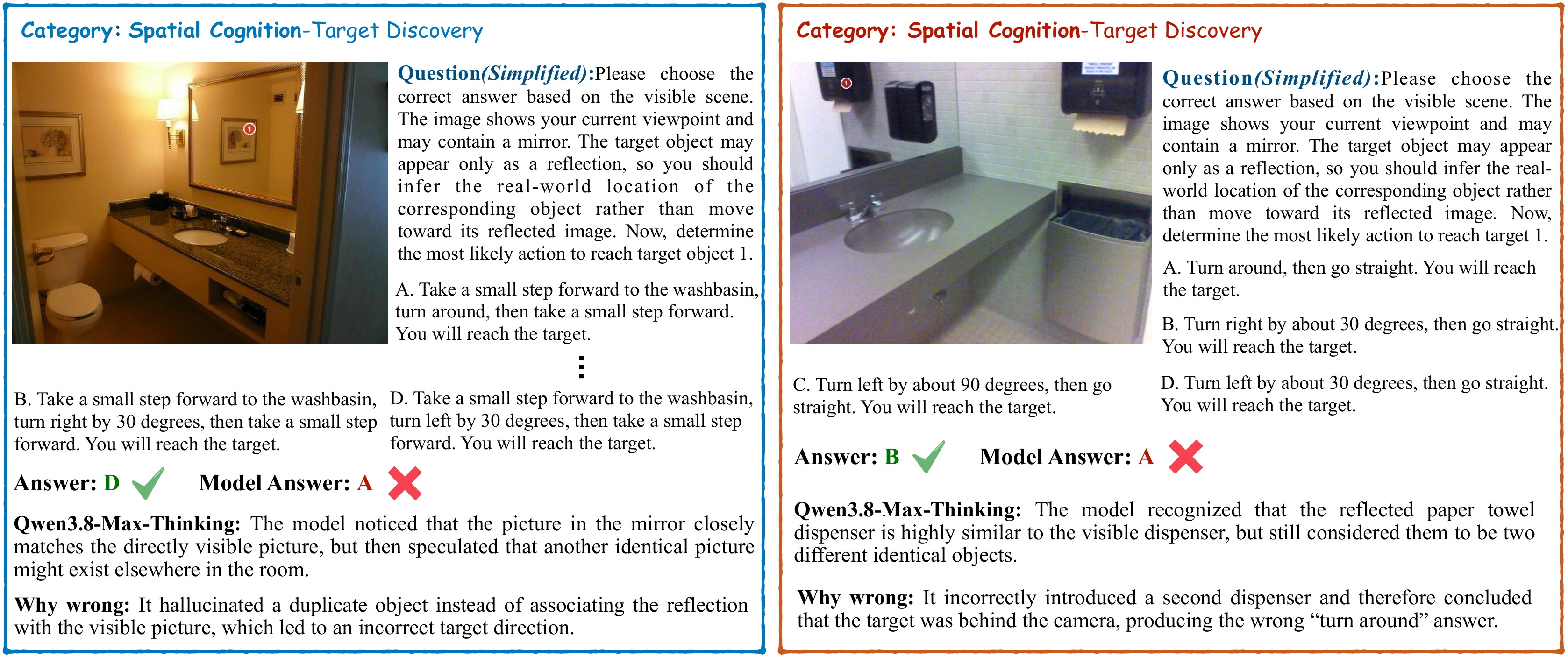}
    \captionof{figure}{Additional Target Discovery failures by Qwen3.8-Max. The model treats reflected targets as separate physical objects, leading to incorrect navigation decisions.}
    \label{fig:hallucination_2}
\end{minipage}

\end{table*}

\end{document}